\documentclass[journal]{IEEEtran}
\usepackage{rotating}
\usepackage{multirow}
\usepackage{ragged2e}  
\usepackage{array}
\usepackage{makecell}
\usepackage{textcomp}
\usepackage{booktabs}
\usepackage{amsfonts}
\usepackage{subfig}  
\usepackage{cite}
\usepackage{dblfloatfix}   
\usepackage{placeins}     

\usepackage{amsmath}
\usepackage{float}   
\usepackage{amssymb}
\usepackage[final]{pdfpages}
\usepackage{nicefrac}
\DeclareMathOperator*{\argmin}{argmin}
\usepackage{dsfont}
\newcommand{\ee}[1]{\mathds{E}[#1]}
\newcommand{\eek}[1]{\mathds{E}_k[#1]}
\newcommand{\agg}[1]{{\mbox{\bfseries AGG}(#1)}}
\newcommand{\bragg}[1]{{\mbox{\bfseries BR-AGG}(#1)}}
\newcommand{\norm}[1]{\|#1\|}
\newcommand{\NORM}[1]{\Big\|#1 \Big\|}
\newcommand{\inp}[1]{\langle #1 \rangle}
\newcommand{\sq}[1]{\Big[ #1 \Big]}
\newcommand{\bb}{Byrd}
\newtheorem{theorem}{\textbf{Theorem}}
\newtheorem{lemma}{\textbf{Lemma}}
\newtheorem{assump}{\textbf{Assumption}}
\newtheorem{remark}{\textbf{Remark}}
\newtheorem{corollary}{\textbf{Corollary}}
\newtheorem{proposition}{\textbf{Proposition}}
\usepackage{algorithm}
\usepackage{algorithmic}
\usepackage{graphicx}
\usepackage{xr}

\begin{document}
	\title{A Byzantine-Robust Federated Learning with Nesterov's Momentum}
	\author{\mbox{Lihan Xu}, 
		\mbox{Xiaoyi Fan},
		\mbox{Gang Wang}, 
		\mbox{Runhao Zeng}, 
		\mbox{Xiping Hu}, and \mbox{Yanjie Dong}
		\thanks{This work was supported by the Joint Laboratory of Intelligent Cyber-Physical Systems 250201013. \emph{(Corresponding author: Y. Dong ydong@xjtu.edu.cn)}.}
		\thanks{L. Xu and Y. Dong are with School of Cyber Science and Engineering, Xi'an Jiaotong University, Xi'an 710049, China.}
		\thanks{X. Fan, R. Zeng, and X. Hu are with the Artificial Intelligence Research Institute and Guangdong-Hong Kong-Macao Joint Laboratory for Emotional Intelligence and Pervasive Computing, Shenzhen MSU-BIT University, Shenzhen 518172, China. }
		\thanks{G. Wang is with the State Key Laboratory of Autonomous
			Intelligent Unmanned Systems, Beijing Institute of Technology, Beijing
			100081, China.	}
		
		}
	\maketitle
	\begin{abstract}
We investigate robust federated learning, where a group of workers collaboratively train a shared model under the orchestration of a central server in the presence of Byzantine adversaries capable of arbitrary and potentially malicious behaviors. 
To improve communication efficiency while retaining resilience against such adversaries, we propose a Byzantine-resilient Nesterov-augmented federated learning (Byrd-NAFL) algorithm.
Byrd-NAFL applies Nesterov's momentum at the server to the output of any soft $\gamma$-Byzantine resilient aggregation rule, while each worker only computes and uploads a stochastic gradient, which keeps the workers stateless, as desired for intermittently participating Internet of Things devices.
For smooth non-convex losses, we establish a finite-time convergence guarantee for all momentum factors $\beta\in[0,1)$ without requiring the aggregated gradient to be unbiased or bounded; the bound reduces to that of Byzantine-resilient stochastic gradient descent at $\beta=0$ and is tighter than that of server-side heavy-ball momentum.
Experiments on MNIST, CIFAR-10, and Sentiment140 show that Byrd-NAFL converges faster and attains higher accuracy than existing benchmarks in most i.i.d. and non-i.i.d. setups under diverse attacks, while its resilience is bounded by that of the underlying aggregation rule.
\end{abstract}
\begin{IEEEkeywords}
Federated learning, Byzantine resilience, Nesterov's momentum, communication efficiency, Internet of Things.
\end{IEEEkeywords}

\section{Introduction}
As a promising paradigm for privacy-preserving distributed learning, federated learning (FL) leverages the computational capabilities of user terminals to learn from decentralized data under the orchestration of a central server.
Since each device only performs local computation and exchanges parameters with the server, FL is naturally suited to large-scale Internet-of-Things (IoT) systems with energy-constrained edge devices and private data~\cite{Ilhan2023,Dong2025,Zhang2026,Liu2026}, e.g., in healthcare~\cite{xu2022healthcare, Shangguan2025}, mobile edge computing~\cite{Ilhan2023, Dong2025}, and autonomous driving~\cite{Lin2024, Fu2024}.
Despite its privacy benefits, vanilla FL~\cite{KonecnyMRR16, McMahan2017} faces two major challenges, namely, Byzantine attacks~\cite{Lamport1982, blanchard2017machine} and the communication bottleneck~\cite{kairouz2021advances}.
Byzantine adversaries may upload arbitrary malicious information to the server~\cite{blanchard2017machine, Zhu2022BRFL,li2019rsa,Zhang2025}, which has motivated various Byzantine-resilient aggregation rules~\cite{blanchard2017machine,chen2017distributed,yin2018byzantine,mhamdi2018bulyan,crsa,rang}.
For example, Krum~\cite{blanchard2017machine} selects the gradient with the smallest sum of squared distances to its nearest neighbors.

The communication bottleneck arises when all workers simultaneously transmit their local updates to the server~\cite{kairouz2021advances}.
In IoT systems, edge devices are often connected to the server through bandwidth-limited wireless channels~\cite{Ilhan2023}, so that the frequent exchange of model parameters can make training prohibitively slow~\cite{kairouz2021advances}.
Current communication-efficient FL algorithms can be classified into three categories: 
(i) reduction of communication frequency~\cite{chen2018lag, stich2019local, Yu2019, lin2020don, Chen2021a,Zhang2025,pmlr-v267-lu25u,zuo2025,rang}, 
(ii) compression of transmitted information~\cite{Mao2022, Chen2021, zheng2024fedcstq, Dong2025,crsa}, and 
(iii) reduction of the total number of iterations~\cite{Yu2019, yang2020fednag, reddi2021adaptive, Wu2023}.

{
Combining Byzantine resilience with iteration reduction is nontrivial, since momentum can accumulate the errors of inexact gradients~\cite{AllenZhu2017Katyusha}, such as the outputs of Byzantine-resilient aggregation rules.
Existing Byzantine-resilient momentum methods keep the momentum at the workers to reduce the variance of the honest inputs~\cite{karimireddy2021learning,el2021distributed,Farhadkhani2022Resam,byz-adavr}, which requires every worker to keep a momentum state across rounds and relies on an independent and identically distributed (i.i.d.) assumption~\cite{allouah2023fixingmixingrecipeoptimal}.
Therefore, we apply Nesterov's momentum~\cite{nesterov1983method} at the server to the output of a Byzantine-resilient aggregation rule, which keeps the workers stateless, as desired for intermittently participating IoT devices, and aims to reduce the number of communication rounds.
Momentum is also known to benefit non-i.i.d. FL by removing the need for a bounded data-heterogeneity assumption~\cite{Cheng}.}

\subsection{Related Works} 
\paragraph{Byzantine resilience}
Byzantine adversaries~\cite{Lamport1982} model both faulty and malicious workers~\cite{blanchard2017machine, 8786075}: they behave arbitrarily, are indistinguishable from honest workers, and are commonly assumed to be fewer than the honest workers.
Since the seminal work~\cite{blanchard2017machine}, the standard mean has been replaced by more resilient aggregation rules, e.g., Krum~\cite{blanchard2017machine}, GeoMed~\cite{chen2017distributed}, CwMed and TrimMean~\cite{yin2018byzantine}, Bulyan~\cite{mhamdi2018bulyan}, RSA~\cite{li2019rsa}, centered clipping (CC)~\cite{karimireddy2021learning}, BRFLATA~\cite{LiLWW25}, H+~\cite{zuo2025}, and ExpGuard~\cite{Christophe2025}.
They exclude statistical outliers~\cite{blanchard2017machine,yin2018byzantine,mhamdi2018bulyan,chen2017distributed}, assign learning- or similarity-based weights~\cite{Christophe2025,LiLWW25,zuo2025}, or suppress the magnitude of malicious updates~\cite{li2019rsa,karimireddy2021learning,Christophe2025}.
{Nevertheless, the resilience of Krum and Bulyan is established for honest gradients that are i.i.d. unbiased estimates of the global gradient, and only their asymptotic convergence is proved~\cite{blanchard2017machine, mhamdi2018bulyan}.}
Moreover, finite-time guarantees are either missing~\cite{zuo2025,Christophe2025} or derived for stochastic gradient descent (SGD)-type updates without momentum, mostly under strongly convex losses~\cite{chen2017distributed, yin2018byzantine, li2019rsa,xia2025}.
In addition, complex strategies entail high communication and computational overhead~\cite{Christophe2025,LiLWW25}.

Worker-side momentum reduces the variance of the honest inputs to robust aggregation~\cite{karimireddy2021learning,el2021distributed,Farhadkhani2022Resam,byz-adavr}.
For instance, D-SHB~\cite{el2021distributed} applies heavy-ball momentum at the workers, proves the reduction of the variance-to-norm ratio, and empirically evaluates Nesterov's momentum; \cite{Farhadkhani2022Resam} derives the convergence of D-SHB without analyzing Nesterov's momentum; and Byz-AdaVR-EF21~\cite{byz-adavr} employs Adam-type momentum and empirically demonstrates its Byzantine resilience.
{Worker-side momentum also counters attacks coupled across rounds, which defeat aggregation rules that ignore the history, such as Krum~\cite{karimireddy2021learning}.}
Nevertheless, these results rely on an i.i.d. assumption: in non-i.i.d. setups, heterogeneous gradients cause the worker-side momentum vectors to drift apart, which degrades or even breaks robust aggregation~\cite{allouah2023fixingmixingrecipeoptimal}.
Moreover, the convergence of Nesterov's momentum under Byzantine-resilient aggregation has yet to be characterized.

\paragraph{Communication efficiency}
For communication frequency reduction, the communication interval is determined by, e.g., adaptive scheduling based on local gradient variations~\cite{chen2018lag} or fixed periodic communication~\cite{stich2019local, lin2020don, Chen2021a,pmlr-v267-lu25u,rang}.
For example, RAGE-Local-SGD~\cite{rang} employs a fixed interval and maintains Byzantine resilience based on covariance.
Compression methods reduce the volume transmitted per round through sparsification~\cite{crsa}, quantization~\cite{Mao2022, Chen2021,pmlr-v267-lu25u}, compressed sensing~\cite{zheng2024fedcstq}, or autoencoder-based encoding~\cite{Dong2025}.
For instance, FedSMU~\cite{pmlr-v267-lu25u} transmits only the signs of the updates, and the Byzantine-resilient C-RSA~\cite{crsa} combines rand-$l$ sparsification with element-wise sign functions.
Iteration-reduction methods for vanilla FL typically exploit momentum-based acceleration~\cite{Yu2019, yang2020fednag, reddi2021adaptive, Wu2023}, whereas server-side momentum has not been analyzed under Byzantine-resilient aggregation.
{Byrd-NAFL belongs to this category: it neither compresses nor skips communication but aims to reach a target accuracy in fewer communication rounds, and is thus complementary to the other two categories.}

\subsection{Our Contributions}
Early research focuses on Byzantine-resilient FL under strongly convex losses and the unbiased gradient aggregation~\cite{blanchard2017machine, mhamdi2018bulyan, chen2017distributed, yin2018byzantine, li2019rsa}. 
	Recent works \cite{crsa, rang} establish convergence guarantees for non-convex Byzantine-resilient FL; however, the theoretical analyses do not incorporate momentum mechanisms. 
	Therefore, the potential benefits of Nesterov's momentum remain theoretically unexplored in the presence of data heterogeneity and adversarial perturbations. 
	Consequently, we are motivated to provide a unified framework that rigorously characterizes the role of momentum in non-convex and Byzantine-resilient optimization.
Different from the aforementioned methods, we aim to develop a momentum-augmented Byzantine-resilient FL algorithm that is applicable to smooth and non-convex loss functions without the unbiasedness assumption~\cite{blanchard2017machine, mhamdi2018bulyan}. 

The integration of momentum into Byzantine-resilient FL algorithms has demonstrated the empirical benefits in Byzantine resilience~\cite{el2021distributed,Farhadkhani2022Resam,byz-adavr}.
Both theoretical and empirical analyses indicate that the convergence of SGD with heavy-ball momentum can deteriorate as the momentum factor approaches one~\cite{shbdw}.
In parallel, Nesterov’s momentum has been empirically evaluated in~\cite{el2021distributed} for Byzantine settings, yet without finite-time convergence guarantees.
Moreover, existing worker-side momentum methods usually suffer from degradation of robust aggregation in non-i.i.d. setups~\cite{allouah2023fixingmixingrecipeoptimal} and the acceleration effect of Nesterov’s momentum remains largely underexplored.

Motivated by the above limitations, we propose to integrate Nesterov’s momentum into a Byzantine-resilient FL framework.
In contrast to existing approaches~\cite{Farhadkhani2022Resam, blanchard2017machine, mhamdi2018bulyan, yin2018byzantine}, our proposed \textbf{\bb-NAFL} algorithm is built on two minimal assumptions:
(i) the Byzantine-resilient aggregation rule returns a search direction that remains close to the true descent direction; and (ii) the loss functions are non-convex but Lipschitz smooth.
Our contributions are as follows.
\begin{itemize}
    \item Our \bb-NAFL algorithm incorporates server-side Nesterov's momentum to enhance communication efficiency in the presence of Byzantine adversaries, while each worker is only required to compute and upload the local stochastic gradient. 
    { In \bb-NAFL, the server-side Nesterov’s momentum avoids the drift of local momentum vectors in non-i.i.d. setups and exploits the aggregated gradient information to improve communication efficiency, while the Byzantine-resilient aggregation rule mitigates malicious updates and preserves resilience against adversarial disruptions.}
    \item We establish the finite-time convergence rate of our proposed \bb-NAFL algorithm for non-convex smooth loss functions under the aforementioned minimal assumptions and quantitatively analyze the impacts of Byzantine adversaries, Nesterov's momentum, and stochastic gradient noise on the learning error.
\end{itemize}

\textbf{Organization.}~The remaining work is organized as follows. 
The problem description and several preliminaries are introduced in Section II.
The algorithmic development and theoretical results are provided in Section III. 
Extensive numerical tests are presented in Section IV, which is followed by the concluding remarks in Section V. 
			
\section{Problem Description and Preliminaries}
\subsection{Vanilla Federated Learning}\label{AA}
We consider an FL system where $N$ workers collaboratively train a global model without uploading the local raw data $\{\mathcal D_n\}_{n=1}^N$ to the central server.
The primary objective of FL is to minimize the global loss $f(x)$ that is defined as the average local loss $f_n(x)$ over the local dataset ${\cal D}_n$ as 
\begin{equation}\label{eq:01}
    \tilde x_{opt} = \argmin_{x \in \mathbb{R}^d} f(x) := 
    \frac{1}{N}\sum_{n=1}^N f_n(x)
\end{equation}
where $x$ is the $d$-dimensional model parameter vector, and 
$f_n(x) := \mathbb{E}_{\xi_n \sim \mathcal{D}_n} [\ell(x; \xi_n)]$ with $\ell(x; \xi_n)$ being the loss evaluated at model $x$ on a mini-batch $\xi_n$ sampled from the local dataset $\mathcal{D}_n$, $n=1, \ldots, N$.

For each communication round, the server broadcasts the current model $x_k$ to all workers, and each worker $n$ can locally compute the stochastic gradient $g_{n,k} = \nabla\ell(x_k; \zeta_{n,k})$ that is an unbiased estimator to $\nabla f_n(x_k)$, i.e., $\nabla f_n(x_k) = \ee{ g_{n,k} }$ with a mini-batch of data uniformly sampled from $\mathcal{D}_n$, $n=1, 2, \ldots, N$. 
The server requires different information for aggregation. 
For instance, the aggregation policy of FedGrad \cite{KonecnyMRR16} per iteration $k$ is
\begin{equation}\label{eq:02}
    \agg{ [g_{n,k}]_{n=1}^N } = \frac{1}{N}\sum_{n=1}^N g_{n,k}
\end{equation}
where $g_{n,k}$ denotes the local gradient per worker $n$. 

\subsection{Byzantine-Resilient Aggregation}
\begin{figure}[htbp]
    \vspace{-0.3 cm}
    \centering
    \includegraphics[width=0.99\linewidth]{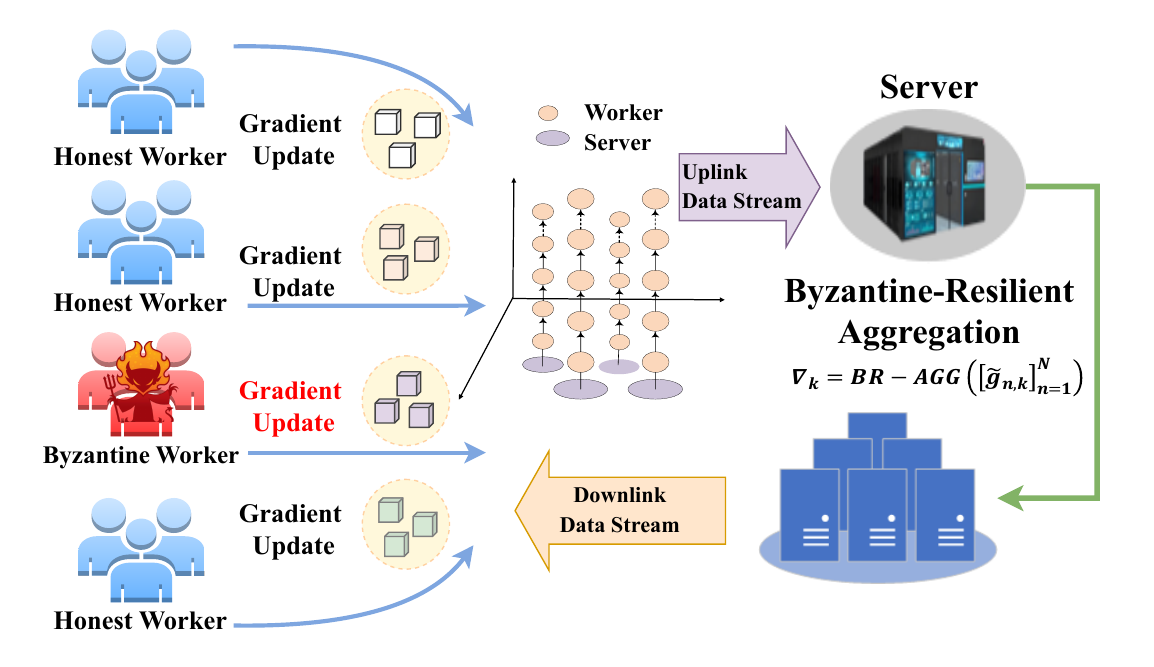}
    \vspace{-0.5 cm}
    \caption{An illustration of Byzantine-resilient aggregation.}
    \label{fig:byrd-agg}
    \vspace{-0.2 cm}
\end{figure}

When a subset of the $N$ workers are Byzantine, the aggregation rule in \eqref{eq:02} fails since Byzantine adversaries may upload arbitrary data to disrupt convergence as shown in Figure~\ref{fig:byrd-agg}.
Without loss of generality, we assume that the first $H$ workers are honest while the remaining $N-H$ workers are Byzantine adversaries.  
Byzantine-resilient aggregation requires honest workers to satisfy $H > N/2$~\cite{Lamport1982}.
The Byzantine-resilient FL minimizes the population loss of the honest workers as 
\begin{equation}\label{eq:04}
    x_{opt} = \argmin_{x \in \mathbb{R}^d} f(x) := 
    \frac{1}{H}\sum_{n=1}^H f_n(x).
\end{equation}

To solve \eqref{eq:04}, each honest worker $n$ with $n = 1, \ldots, H$ follows the prescribed protocol to upload the local gradient $g_{n,k}$ to the server truthfully while each Byzantine worker $n = H+1, \ldots, N$ can upload any crafted information $g_{n,k}^*$ to the server with the intent to disrupt the training process. 
More specifically, the server receives local information (e.g., local gradients and local models) from $N$ workers as 
\begin{equation}\label{eq:05}
    \tilde g_{n,k} =     
    \begin{cases}
        g_{n,k} & n = 1, \ldots, H  \\
        g^*_{n,k} & n = H+1, \ldots, N.
    \end{cases}
\end{equation}

Note that honest workers and Byzantine adversaries simultaneously upload local information as in \eqref{eq:05}, and the server cannot distinguish between malicious and honest training information. 
Therefore, it must adopt a Byzantine-resilient aggregation rule to mitigate the impact of adversarial inputs. 
Based on \eqref{eq:05}, the Byzantine-resilient aggregation rule of local gradients can be defined as
\begin{equation}\label{eq:06}
    \nabla_k = \bragg{ [\tilde g_{n,k}]_{n=1}^N }. 
\end{equation}

To enable Byzantine-resilient FL algorithms, the aggregation rule in \eqref{eq:06} should satisfy the following assumption \cite{blanchard2017machine, chen2017distributed, mhamdi2018bulyan} to guarantee the aggregated gradient is not far away from the ground-truth gradient $\nabla f(x_k) = \frac{1}{H}\sum_{n=1}^{H} \ee{ g_{n,k} } $ per recursion $k$.
In other words, the aggregation rule in \eqref{eq:06} is called soft $\gamma$-Byzantine resilient when the following assumption is satisfied. 

\begin{assump}[Soft $\gamma$-Byzantine resilient]\label{as:01}
    The aggregation rule $\nabla_k = \mbox{\bfseries BR-AGG}( [\tilde g_{n,k}]_{n=1}^N )$ is soft  $\gamma$-Byzantine resilient when the output $\nabla_k$ per recursion $k$ satisfies
    \begin{align}
        \inp{ \eek{\nabla_k }, \nabla f(x_k) } &\ge (1-\sin\gamma) \norm{\nabla f(x_k)}^2 \label{eq:07}\\
        \eek{ \norm{\nabla_k}^2 } &\le c_1 \norm{\nabla f(x_k)}^2 + c_2 \label{eq:08}
    \end{align}
    where $\eek{\cdot}$ denotes the conditional expectation with respect to the natural filtration up to iteration $k$, $\gamma \in (0, \frac{\pi}{2})$, and $c_1$ and $c_2$ are positive constants.
\end{assump}

The Byzantine-resilient hyperparameters $\gamma, c_1$, and $c_2$ in Assumption~\ref{as:01} possess clear physical meanings. 
Specifically, $\gamma$ represents the maximum angle between the aggregated gradient and the true gradient. 
The constants $c_1$ and $c_2$ indicate the magnitude scaling and the error floor from local gradient variance and malicious attacks. 
\textcolor{black}{
Therefore, eq.~\eqref{eq:07} requires the aggregated
gradient to preserve a positive alignment with the ground-truth gradient and ensure a valid descent direction. 
In addition, eq.~\eqref{eq:08} imposes an upper bound on the second-order moment of the aggregated gradient to prevent an unbounded update magnitude.
Collectively, Assumption~\ref{as:01} ensures that the Byzantine-resilient aggregator yields a reliable descent direction while bounding the magnitude and variability of the aggregated gradient even under adversarial perturbations.
}

Compared to the conventional $\gamma$-Byzantine resilience assumption~\cite{blanchard2017machine, chen2017distributed, mhamdi2018bulyan}, \emph{the soft $\gamma$-Byzantine resilience condition relaxes the stringent requirement of unbiasedness on the aggregated gradient $\nabla_k$ and imposes weaker constraints on its second-order moment.}
Therefore, the soft $\gamma$-Byzantine resilience can find broader applicability in adversarial FL scenarios.
Moreover, Assumption~\ref{as:01} can be satisfied by several well-established aggregation rules, such as Krum~\cite{blanchard2017machine}, CwMed~\cite{yin2018byzantine}, Bulyan~\cite{mhamdi2018bulyan}, and GeoMed~\cite{chen2017distributed}. 
For instance,~\cite[Proposition 1]{blanchard2017machine} formally verifies that the inequalities~\eqref{eq:07} and~\eqref{eq:08} are satisfied under the Krum aggregation rule. 
Similarly,~Lemma 2.1~in \cite{chen2017distributed} establishes that the GeoMed aggregation rule also satisfies Assumption \ref{as:01}.

In practice, constructing statistically valid confidence intervals for $(\gamma,c_1,c_2)$ is challenging due to their dependence on the unknown true gradient norm $\|\nabla f(x_k)\|$, the underlying data distribution, and the type and strength of potentially adaptive Byzantine attacks. 
However, the above parameters determine the admissible upper bound of the stepsize and the effective range of the momentum factor, rather than directly influencing the algorithm (discussed later in Section~\ref{conv-anal}).
Consequently, rather than directly estimating $(\gamma, c_1, c_2)$, we can select different stepsizes and momentum coefficients for different aggregation methods through ablation studies to ensure the convergence of the algorithm.
{Nevertheless, Appendix~\ref{app:constants} in the supplementary materials shows how $(\gamma,c_1,c_2)$ can be bounded in closed form for robust aggregation rules and estimated in simulation.}

\subsection{Revisiting Nesterov's Momentum}
First proposed by Yurii Nesterov, Nesterov's momentum can leverage the historical gradient information to augment and stabilize the convergence by using a mini-batch of samples per recursion~\cite{nesterov1983method}. 
The standard Nesterov's momentum recursion can be cast as~\cite{nesterov1983method, YuanOct.2016}
\begin{subequations}\label{eq:09}
    \begin{align}
        y_k &= x_k - \eta \nabla f(x_k) \label{eq:09a}\\
        x_{k+1} &= y_k + \beta(y_k - y_{k-1}) \label{eq:09b}
    \end{align}
\end{subequations}
where $\eta$ is the stepsize and $\beta$ is the momentum factor. 

Substituting \eqref{eq:09a} into \eqref{eq:09b} and dividing both sides by $\eta\beta$, we obtain
\begin{subequations}\label{eq:10}
    \begin{align}
        z_{k+1} &= \beta z_{k} + \nabla f(x_k) \\
        \eta\beta z_{k+1} &= x_k - x_{k+1} - \eta\nabla f(x_k).
    \end{align}
\end{subequations}

Setting $y_k = \beta z_{k+1} + \nabla f(x_k)$, we obtain $x_{k+1} = x_k - \eta y_k$. Therefore, the standard Nesterov's momentum recursion can be recast as 
\begin{subequations}\label{eq:11}
    \begin{align}
        z_{k+1} &= \beta z_k +  \nabla f(x_k) \label{eq:11a}\\
        y_{k} &= \beta z_{k+1} +  \nabla f(x_k) \label{eq:11b} \\
        x_{k+1} &= x_k - \eta y_k. \label{eq:11c}
    \end{align}
\end{subequations}

Based on \eqref{eq:11} with $z_0 = 0$, we obtain
\begin{equation}\label{eq:12}
    \frac{1}{\eta}(x_k \!-\! x_{k+1}) = (1+\beta)\nabla f(x_k) \!+\! \sum_{t=0}^{k-1} \beta^{k+1-t} \nabla f(x_t)
\end{equation}
where the two terms on the right-hand side (RHS) show that Nesterov’s momentum augments convergence by leveraging historical gradients to refine the current search direction.

\section{Byrd-NAFL and Its Convergence Analysis}
In this section, we explore the way to combine Nesterov's momentum with the Byzantine-resilient aggregation rules in the adversarial scenarios and propose the Byrd-NAFL algorithm. 
Then, we characterize the convergence behavior via a finite-time analysis to illustrate the impacts of Byzantine adversaries and Nesterov's momentum factor. 

\begin{algorithm}[ht]
    \caption{Byzantine-Resilient Nesterov-Augmented FL}
    \label{alg:byzantine-katyusha}
    \begin{algorithmic}[1]
        \REQUIRE datasets for honest workers $\{\mathcal{D}_n\}_{n=1}^H$; 
        stepsize $\eta$; momentum factor $\beta$; number of iterations $K$.
        \STATE Server initializes $z_0 = 0$ and initial model parameter $x_0$
        \FOR{$k = 0, 1, \ldots, K - 1$}
        \STATE Server broadcasts global model $x_k$ to all workers
        \STATE Each honest worker samples a mini-batch of samples $\zeta_{n,k}$ and calculates the local stochastic gradient $g_{n,k} = \nabla\ell(x_k; \zeta_{n,k})$, $n = 1, \ldots, H$
        \STATE Each worker uploads the local information to the server \label{alg1:upload}
        \begin{equation}\label{eq:13}
            \tilde g_{n,k} =     
            \begin{cases}
                g_{n,k} & n = 1, \ldots, H  \\
                g^*_{n,k} & n = H+1, \ldots, N
            \end{cases}
        \end{equation}
        \STATE Server aggregates local stochastic gradients as \label{alg1:agg}
        \begin{equation}\label{eq:14}
            \nabla_k = \bragg{ [\tilde g_{n,k}]_{n=1}^N }
        \end{equation}
        \STATE Server updates the global model as \label{alg1:model-update}
        \begin{equation}\label{eq:15}
            \begin{split}
                z_{k+1} &= \beta z_k +  \nabla_k \\
                y_{k} &= \beta z_{k+1} +  \nabla_k \\
                x_{k+1} &= x_k - \eta y_k 
            \end{split}
        \end{equation} 
        \ENDFOR
     \RETURN $x_K$
    \end{algorithmic}
\end{algorithm}

\vspace{-0.5 cm}
\subsection{Byrd-NAFL Algorithm}\label{sec:3.B}
In the previous section, we demonstrated that Nesterov’s momentum augments and stabilizes convergence by incorporating historical gradient information. 
Motivated by this insight, we propose to leverage historical Byzantine-resilient gradients to develop Byrd-NAFL to operate robustly in the presence of Byzantine adversaries.
The detailed procedures of the Byrd-NAFL are summarized in Algorithm \ref{alg:byzantine-katyusha}.
\begin{itemize}
    \item On the worker side, each worker $n$ maintains a local copy of the global model $x_k$. 
    At each iteration $k$, each honest worker $n$ uniformly samples a mini-batch $\zeta_{n,k}$ from the local dataset $\mathcal{D}_n$, and computes the corresponding local stochastic gradient $g_{n,k}$.
    At step \ref{alg1:upload} of the Byrd-NAFL algorithm, all workers upload their local information to the server, where honest workers upload the local stochastic gradients $[g_{n,k}]_{n=1}^H$, while Byzantine adversaries may upload arbitrary misleading information.
    \item The server is responsible for the Byzantine-resilient aggregation and model updates per recursion $k$.
    After receiving local information from both honest workers and Byzantine adversaries, the server performs the aggregation to collect a Byzantine-resilient gradient $\nabla_k$ at step \ref{alg1:agg} of the Byrd-NAFL algorithm.
    The Byzantine-resilient gradient $\nabla_k$ is then used to update the global model at step \ref{alg1:model-update} of the Byrd-NAFL algorithm.
\end{itemize}

Overall, each worker computes a local stochastic gradient and uploads it to the server. 
Honest workers send true gradients, while Byzantine adversaries may send arbitrary misleading information.
The server applies a Byzantine-resilient aggregation to obtain a robust gradient, which is then used to update the global model via Nesterov's momentum. Such a process repeats each iteration to ensure reliable convergence.

 In Byrd-NAFL, a Byzantine-resilient aggregation rule is first applied, followed by the computation of the momentum term. 
{
The aggregation rule mitigates the impact of malicious updates, while the momentum term leverages current and historical aggregated gradients to improve convergence.}
In the subsequent subsection, we establish the theoretical convergence guarantees of the proposed Byrd-NAFL algorithm for smooth and non-convex loss functions and quantitatively analyze the impacts of Byzantine adversaries, Nesterov’s momentum, and
stochastic gradient noise on the learning error.

\subsection{Theoretical Results}
\label{conv-anal}
Hereinafter, we will establish our major theoretical results based on the following smoothness assumption.

\begin{assump}[$L$-Lipschitz Smoothness]\label{as:02}
    The loss function $f$ has an $L$-Lipschitz continuous gradient{, i.e., $\norm{\nabla f(x') - \nabla f(x)} \le L\norm{x' - x}$, which implies}
    \begin{equation}\label{eq:16}
        f(x') - f(x) \le \inp{\nabla f(x), x' - x} + \frac{1}{2}L\norm{x' - x}^2
    \end{equation}
    where $x'$ and $x$ are both $d$-dimensional vectors.
\end{assump}

Assumption~\ref{as:02} holds for common losses, e.g., the squared $\ell_2$-norm and the (multinomial) logistic regression loss, and for certain classes of neural networks~\cite{Latorre}.
Under Assumptions~\ref{as:01} and~\ref{as:02}, we establish the convergence of Byrd-NAFL in Theorem~\ref{thm:01} and that of its server-side heavy-ball (SHB) counterpart in Theorem~\ref{thm:02}{, whose proofs are provided in Appendices~\ref{app:proof-thm1} and~\ref{app:shb_cov}, respectively}.

{
\begin{theorem}\label{thm:01}
    Suppose Assumptions \ref{as:01} and \ref{as:02} are satisfied, $0\le\beta<1$, and $\Delta = \ee{f(x_0)}- f^*$ with $f^* = \inf_x f(x) > -\infty$.
    Let $\rho := 1-\sin\gamma$ and $\phi_\beta := \nicefrac{(2\beta^2-\beta+1)}{(1-\beta)}$.
    When the stepsize satisfies $\eta \le \nicefrac{(1-\beta)\rho}{(L c_1 \phi_\beta)}$, the convergence rate of the Byrd-NAFL algorithm is
    \begin{equation}\label{eq:18}
    \frac{1}{K}\sum_{k=0}^{K-1} \ee{ \norm{\nabla f(x_k)}^2 }
    \le \frac{2(1-\beta)\Delta}{\rho \eta K} 
    + \frac{L\phi_\beta c_2}{(1-\beta)\rho} \eta.
    \end{equation}
\end{theorem}}

{
When $\beta=0$, Byrd-NAFL reduces to Byzantine-resilient SGD. Since $\phi_0=1$, Theorem~\ref{thm:01} then gives, for $0<\eta\le\nicefrac{\rho}{(L c_1)}$,
\begin{equation}
	\frac{1}{K}\sum_{k=0}^{K-1}\ee{\norm{\nabla f(x_k)}^2}
	\le \frac{2\Delta}{\rho\,\eta K} 
	+ \frac{L c_2}{\rho}\,\eta.
	\label{eq:beta-zero-bound}
\end{equation}

\label{par:thm1-disc}%
Theorem~\ref{thm:01} covers the whole range $\beta\in[0,1)$ for momentum factor, and~\eqref{eq:beta-zero-bound} is the baseline (i.e., Byzantine-resilient SGD) against which the effect of the momentum factor is assessed.
Note that the convergence rate of the Byrd-NAFL in Theorem~\ref{thm:01} has the same form as the Byzantine-resilient SGD~\eqref{eq:beta-zero-bound}, with the stepsize $\eta$ replaced by the effective stepsize $\nicefrac{\eta}{(1-\beta)}$ (cf. Proposition~\ref{lem:equal-effective-stepsize}) and the smoothness constant $L$ inflated by the momentum-induced factor $\phi_\beta\ge 1$.
More specifically, compared with Byzantine-resilient SGD under the same stepsize, Byrd-NAFL reduces the initialization-dependent term by the factor $1-\beta$ at the expense of enlarging the error floor by the effective stepsize $\nicefrac{\eta}{(1-\beta)}$ and the momentum-induced factor $\phi_\beta$.

\label{par:roles}%
Theorem~\ref{thm:01} also separates the roles of the two components of Byrd-NAFL: the aggregation rule enters only through $(\gamma, c_1, c_2)$ in Assumption~\ref{as:01}, e.g., a larger $\gamma$ enlarges the initialization-dependent term and the error floor and shrinks the stepsize bound, whereas the momentum enters only through $\nicefrac{\eta}{(1-\beta)}$ and the momentum-induced factor $\phi_\beta$.
Hence, the momentum mechanism cannot lower the error floor that is determined by the aggregation rule, and the resilience of Byrd-NAFL is inherited from its aggregation rule; 
in particular, the Byzantine resilience vanishes as $\rho\to0$, i.e., once the aggregated gradient loses its positive alignment with $\nabla f(x_k)$ in~\eqref{eq:07}.
We will also verify the merits of Byrd-NAFL empirically in Sections~\ref{sec:4.B} and~\ref{sec:4.C} of the numerical tests.

}

{
For comparison, SHB replaces $y_k$ in the third equation of~\eqref{eq:15} by $z_{k+1}$, i.e., $x_{k+1} = x_k - \eta z_{k+1}$.

\begin{theorem}\label{thm:02}
    Suppose Assumptions \ref{as:01} and \ref{as:02} are satisfied and $0\le\beta<1$; let $\Delta$ and $\rho$ be defined as in Theorem~\ref{thm:01}, and let $\phi_\beta^{\mbox{\tiny HB}} := \nicefrac{(1+\beta)}{(1-\beta)}$.
    When the stepsize satisfies $\eta \le \nicefrac{(1-\beta)\rho}{(L c_1\phi_\beta^{\mbox{\tiny HB}})}$, the convergence rate of SHB is
    \begin{equation}\label{eq_r2:hb-final-bound1}
            \frac{1}{K}\sum_{k=0}^{K-1} \ee{\norm{\nabla f(x_k)}^2}
            \le \frac{2(1-\beta)\Delta}{\rho\,\eta K} + \frac{L\phi_\beta^{\mbox{\tiny HB}} c_2}{(1-\beta)\rho}\,\eta.
\end{equation}
\end{theorem}

To gain deeper insight into the advantages of Byrd-NAFL, we characterize the effective stepsizes of Byrd-NAFL and SHB near convergence in Proposition~\ref{lem:equal-effective-stepsize} with the proof deferred to Appendix~\ref{app:acc}.

\begin{proposition}[Effective stepsize near convergence]
	\label{lem:equal-effective-stepsize}
	Suppose that, near convergence, the outputs of the aggregation rule remain constant for all relevant iterations, i.e., $\nabla_t = g$ for $t=0,\ldots,k$.
	Then $x_{k+1} = x_k - \eta_{\mathrm{eff},k}\, g$, where $\eta_{\mathrm{eff},k} = \eta(1-\beta^{k+2})/(1-\beta)$ for Byrd-NAFL and $\eta^{\mbox{\tiny HB}}_{\mathrm{eff},k} = \eta(1-\beta^{k+1})/(1-\beta)$ for SHB.
	Hence, the effective stepsize of Byrd-NAFL exceeds that of SHB by $\eta\beta^{k+1}$, and both tend to $\nicefrac{\eta}{(1-\beta)}$ as $k\to\infty$.
\end{proposition}

Based on Theorem~\ref{thm:01}, Theorem~\ref{thm:02}, and Proposition~\ref{lem:equal-effective-stepsize}, we can now compare Byrd-NAFL with Byzantine-resilient SGD and SHB under the same stepsize $\eta$.
For Byzantine-resilient SGD in~\eqref{eq:beta-zero-bound}, the stepsize $\eta$ alone governs the trade-off between the initialization-dependent term and the error floor (i.e., the first and second terms in~\eqref{eq:18}--\eqref{eq_r2:hb-final-bound1}, respectively).
The momentum introduces a second degree of freedom.
More specifically, through the effective stepsize $\nicefrac{\eta}{(1-\beta)}$, Byrd-NAFL and SHB reduce the initialization-dependent term by the factor $1-\beta$ over that of SGD, at the expense of an error floor enlarged by the factors $\nicefrac{\phi_\beta}{(1-\beta)}$ and $\nicefrac{\phi_\beta^{\mbox{\tiny HB}}}{(1-\beta)}$, respectively, both of which increase with $\beta$.

Between the two Byrd-NAFL and SHB algorithms, the initialization-dependent terms  and the effective stepsizes coincide as $k$ grows since their difference $\eta\beta^{k+1}$ vanishes geometrically (cf. Proposition \ref{lem:equal-effective-stepsize}).
Hence, the advantage of Byrd-NAFL over SHB lies in the error floor.
Note that $\phi_\beta=\phi_\beta^{\mbox{\tiny HB}}-2\beta$, the error floor term of \mbox{Byrd-NAFL} is $\nicefrac{(2\beta^2-\beta+1)}{(1+\beta)}<1$ times that of SHB for every $\beta\in(0,1)$.
Consequently, Byrd-NAFL exploits the second degree of freedom more efficiently: for the same $\eta$ and error floor, it admits a larger $\beta$, and thus a smaller initialization-dependent term than SHB, which improves the constants of the bound rather than its order in $K$.
Such advantage essentially stems from the search direction rather than the effective stepsize: by~\eqref{eq:15}, Byrd-NAFL moves along $y_k=(1+\beta)\nabla_k+\beta^2 z_k$, whereas SHB moves along $z_{k+1}=\nabla_k+\beta z_k$, where $z_k$ accumulates the past aggregated gradients.
The smaller weight $\beta^2$ on $z_k$ has two effects: it makes the deviation $v_k-x_k$ in~\eqref{eqproof:03} proportional to $\beta^2$ rather than $\beta$ as in~\eqref{eq_r2:hb-deviation}, which yields $\phi_\beta<\phi_\beta^{\mbox{\tiny HB}}$, and it limits the accumulation of malicious components that the aggregation rule fails to remove.

}

{
\begin{remark}[Selection of Byzantine-resilient hyperparameters]\label{rem:hyper}
The Byzantine-resilient hyperparameters $(\gamma,c_1,c_2)$ characterize the aggregation rule under a given attack and data distribution, and they depend on $\eta$ and $\beta$ only through the trajectory $\{x_k\}$.
For $(N-H,\lambda)$-robust aggregation rules, e.g., Krum, GeoMed, and TrimMean, they follow in closed form from the robustness coefficient $\lambda$ of the aggregation rule (cf. Table~\ref{tab:lambda}), the variance $\sigma^2$ of the honest stochastic gradients, and the heterogeneity $\omega^2$ of the honest local gradients: $c_1=2$, $c_2=\nicefrac{2\sigma^2}{H}+2\lambda(\sigma^2+\omega^2)$, and $\sin\gamma=\sqrt{\lambda(\sigma^2+\omega^2)}/\nu$ at the iterates with $\norm{\nabla f(x_k)}\ge\nu$ for any $\nu>\sqrt{\lambda(\sigma^2+\omega^2)}$ (Lemma~\ref{lem:robust-constants} in Appendix~\ref{app:constants}).
Since~\eqref{eq:07} may fail closer to stationarity, Lemma~\ref{lem:robust-constants} also provides a relaxed version of~\eqref{eq:07} that holds at every iterate, under which Theorem~\ref{thm:01} remains valid with $\rho=\nicefrac{1}{2}$, i.e., for $\eta\le\nicefrac{(1-\beta)}{(4L\phi_\beta)}$, up to an additional non-vanishing term $2\lambda(\sigma^2+\omega^2)$ (Corollary~\ref{cor:relaxed}).
For aggregation rules without a finite $\lambda$, e.g., CC, $(\gamma,c_1,c_2)$ can be estimated in simulation by replaying the aggregation step at a few checkpoints along the training trajectory (Appendix~\ref{app:constants-sim}).
In all cases, a larger $\lambda$ or a stronger attack decreases $\rho$ or increases $c_1$ or $c_2$, which tightens the stepsize condition of Theorem~\ref{thm:01} and raises its error floor; hence, $\eta$ and $\beta$ should be tuned for the aggregation rule in use under the most severe anticipated attack, as in Section~\ref{sec:4.A}3.
\end{remark}
}

\section{Numerical Results}
\label{exp}

\subsection{Experimental Setup}\label{sec:4.A}
\subsubsection{\bfseries Datasets and Models}
We evaluate Byrd-NAFL on three popular datasets, i.e., the MNIST~\cite{minst}, CIFAR-10~\cite{cifar}, and Sentiment140~\cite{sentiment} datasets.
To evaluate Byrd-NAFL across different scales of complexity, we assign specific model architectures for each dataset.

\begin{itemize}
	\item \textbf{LeNet-5 for MNIST.}
	For the MNIST dataset, we adopt LeNet-5~\cite{minst} that consists of two convolutional layers (each followed by a max-pooling layer) and three fully connected layers.
	LeNet-5 serves as a standard testbed for evaluating convergence speed and Byzantine resilience.
	\item \textbf{ResNet-18 for CIFAR-10.}
	For the more complex CIFAR-10 dataset, we adopt ResNet-18~\cite{resnet} (i.e., an 18-layer residual network) whose shortcut connections facilitate the training of deep networks.
	\item \textbf{SentimentTransformer for Sentiment140.}
	For the sequential text data of Sentiment140, we design SentimentTransformer that is a lightweight classifier built on the Transformer encoder~\cite{transformer}.
	It consists of a word embedding layer with sinusoidal positional encoding and two encoder layers that have four attention heads and a hidden dimension of 256 per encoder.
\end{itemize}

\subsubsection{\bfseries Federated Settings}
For a fair comparison, all methods share the same system configuration and training budget for each dataset.
{
Unless otherwise specified, we use Krum~\cite{blanchard2017machine} as the Byzantine-resilient aggregation rule, since it shows the most consistent stability among the four aggregation rules compared in Section~\ref{sec:ablation}.}
We denote the fraction of Byzantine workers by $\epsilon := (N-H)/N$ and count the budget in training iterations.
All configurations satisfy the requirement $N > 2(N-H)+2$ of Krum~\cite{blanchard2017machine}.
\begin{itemize}
	\item \textbf{MNIST.}
	We set $N = 500$ with $\epsilon \in \{0.3, 0.4\}$.
	Each honest worker uses a mini-batch size of 50, and training runs for 2,000 steps.
	\item \textbf{CIFAR-10.}
	We set $N = 40$ with $\epsilon \in \{0.2, 0.3\}$.
	Each honest worker uses a mini-batch size of 32, and training is extended to 6,000 steps to ensure convergence.
	\item \textbf{Sentiment140.}
	We set $N = 20$ with $\epsilon = 0.4$ in the i.i.d. setup and $\epsilon = 0.3$ in the non-i.i.d. setup of Section~\ref{sec:4.C}.
	Each honest worker uses a mini-batch size of 32, and training runs for 6,000 steps.
\end{itemize}

\subsubsection{\bfseries Hyperparameter Tuning Protocol}
For each dataset, we tune the stepsize and the momentum coefficients of every compared method under the severe Sign-Flipping (SF) attack~\cite{sf}.
All methods share the same candidate grids: the stepsize is selected from $\{1\times10^{-5}, 1\times10^{-4}, 0.001, 0.005, 0.01, 0.05, 0.1\}$, and the momentum coefficient is from $\{0.8, 0.85, 0.9, 0.95, 0.99, 0.999\}$.
For each method, we select the configuration with the highest test accuracy within the same training budget, and we examine the sensitivity to both hyperparameters in Section~\ref{sec:ablation}.
Table~\ref{tab:hyperparams} summarizes the selected hyperparameters.

\begin{table}[h]
	\centering
	\footnotesize
	\setlength{\tabcolsep}{3.8pt}
	\caption{Hyperparameter settings.}
	\label{tab:hyperparams}
	\begin{tabular}{llcc}
		\toprule
		\textbf{Dataset} & \textbf{Method} &
		\textbf{Stepsize} & \textbf{Momentum} \\
		\midrule
		\multirow{9}{*}{MNIST}
		& SGD            & 0.01  & -- \\
		& SHB            & 0.01  & 0.95 \\
		& SHB-DW         & 0.005  & 0.95 \\
		& D-SHB          & 0.05  & 0.9 \\
		& D-SHB-DW       & 0.05  & 0.9 \\
		& Byz-AdaVR-EF21 & $1\times10^{-5}$  & $\beta_1=0.9,\beta_2=0.999$  \\
		& C-RSA           & 0.005 & -- \\
		& RAGE-Local-SGD  & 0.001 & -- \\
		& \textbf{Byrd-NAFL (ours)} &  0.01  & 0.95 \\
		\midrule
		\multirow{9}{*}{CIFAR-10}
		& SGD             & 0.01 & -- \\
		& SHB             & 0.005 & 0.9 \\
		& SHB-DW          & 0.005 & 0.9 \\
		& D-SHB           & 0.01 & 0.9 \\
		& D-SHB-DW        & 0.01 & 0.9 \\
		& Byz-AdaVR-EF21 &  $1\times10^{-5}$  & $\beta_1=0.9,\beta_2=0.999$  \\
		& C-RSA           & 0.005 & -- \\
		& RAGE-Local-SGD  & 0.001 & -- \\
		& \textbf{Byrd-NAFL (ours)} &  0.005  & 0.95 \\
		\midrule
		\multirow{9}{*}{Sentiment140}
		& SGD             & 0.001 & -- \\
		& SHB             & 0.001 & 0.9 \\
		& SHB-DW          & 0.001 & 0.9 \\
		& D-SHB           & 0.005 & 0.9 \\
		& D-SHB-DW        & 0.001 & 0.85 \\
		& Byz-AdaVR-EF21 &$1\times10^{-5}$  & $\beta_1=0.85,\beta_2=0.9$  \\
		& C-RSA           & 0.001 & -- \\
		& RAGE-Local-SGD  & $1\times10^{-4}$ & -- \\
		& \textbf{Byrd-NAFL (ours)} &  0.005  & 0.85 \\
		\bottomrule
	\end{tabular}
\end{table}

\begin{figure*}[t]
	\centering
	\includegraphics[width=0.95\linewidth]{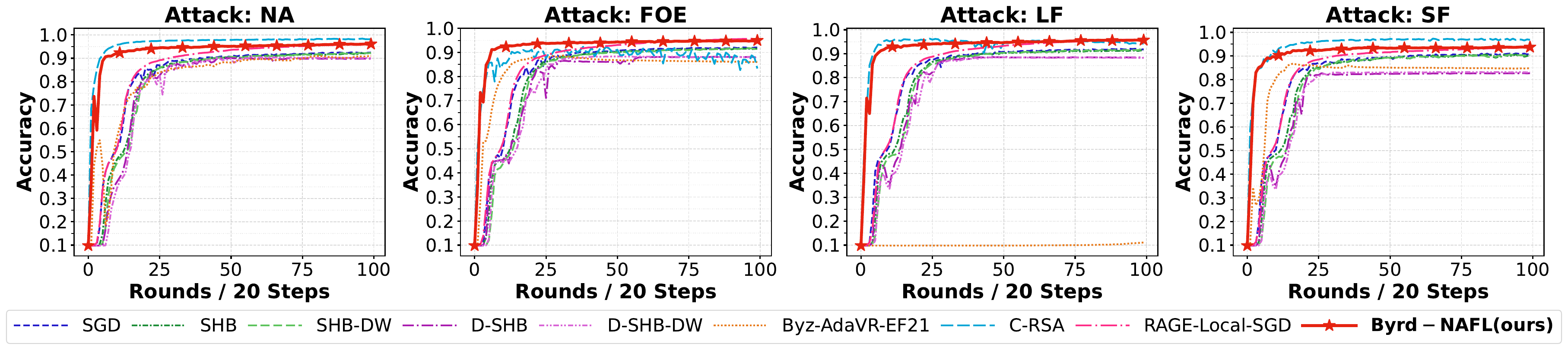}    
	\caption{Test accuracy in the i.i.d. setup ($\epsilon=0.3$) for the MNIST dataset.} 
	\label{fig:opt_03}
	\includegraphics[width=0.95\linewidth]{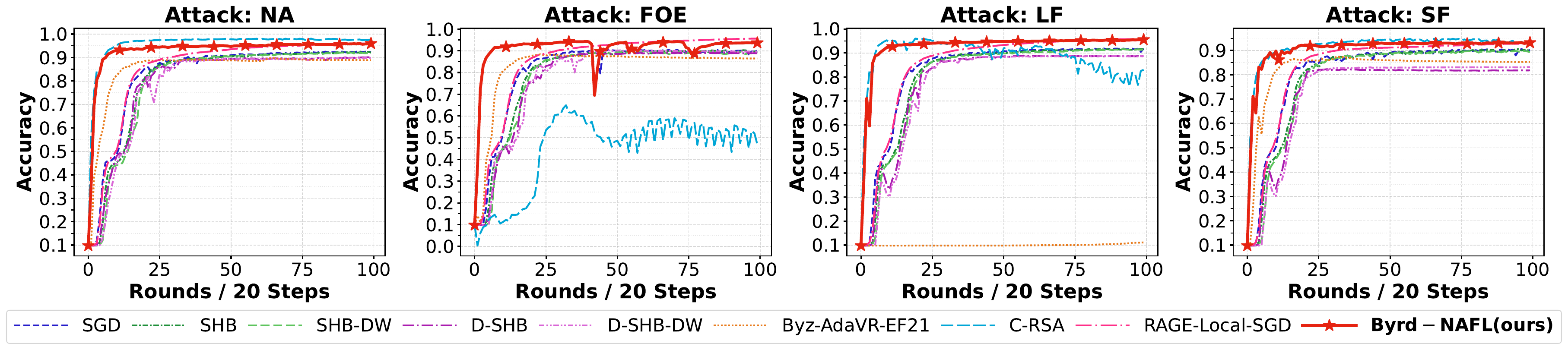}
	\caption{Test accuracy in the i.i.d. setup ($\epsilon=0.4$) for the MNIST dataset.} 
	\label{fig:opt_04}
\end{figure*}

\begin{table*}[t]
	\centering
	\caption{Per-worker communication volume and wall-clock time to reach 80\% of the peak accuracy on MNIST. $>$: target not reached within 2,000 steps; bold: best value per column and metric.}
	\label{tab:communication_mnist}
	\begin{tabular}{lc cccc cccc}
		\toprule
		\multirow{2}{*}{Method} & \multirow{2}{*}{Metric}
		& \multicolumn{4}{c}{MNIST ($\epsilon = 0.3$)}
		& \multicolumn{4}{c}{MNIST ($\epsilon = 0.4$)} \\
		\cmidrule(lr){3-6} \cmidrule(lr){7-10}
		& & NA & FOE & LF & SF & NA & FOE & LF & SF \\
		\midrule
		\multicolumn{2}{l}{Target Accuracy}
		& 79\% & 77\% & 77\% & 78\% & 78\% & 76\% & 77\% & 76\% \\
		\midrule
		\multirow{2}{*}{SGD} & Volume (MB)
		& 80.03 & 75.33 & 75.33 & 75.33
		& 75.33 & 75.33 & 75.33 & 75.33 \\
		& Time (s)
		& 194.5 & 189.8 & 228.2 & 213.1
		& 183.0 & 189.8 & 228.2 & 213.1 \\
		\addlinespace[2pt]
		\multirow{2}{*}{SHB} & Volume (MB)
		& 94.16 & 89.45 & 89.45 & 94.16
		& 94.16 & 89.45 & 89.45 & 89.45 \\
		& Time (s)
		& 287.6 & 271.7 & 298.7 & 300.4
		& 287.6 & 271.7 & 298.7 & 285.4 \\
		\addlinespace[2pt]
		\multirow{2}{*}{SHB-DW} & Volume (MB)
		& 98.86 & 94.16 & 94.16 & 98.86
		& 98.86 & 94.16 & 94.16 & 94.16 \\
		& Time (s)
		& 257.9 & 268.8 & 320.8 & 319.6
		& 257.9 & 268.8 & 320.8 & 304.4 \\
		\addlinespace[2pt]
		\multirow{2}{*}{D-SHB} & Volume (MB)
		& 94.16 & 98.86 & 98.86 & 103.57
		& 94.16 & 108.28 & 103.57 & 98.86 \\
		& Time (s)
		& 220.4 & 245.3 & 286.0 & 284.7
		& 220.4 & 268.6 & 299.6 & 271.7 \\
		\addlinespace[2pt]
		\multirow{2}{*}{D-SHB-DW} & Volume (MB)
		& 98.86 & 112.99 & 108.28 & 108.28
		& 98.86 & 108.28 & 103.57 & 98.86 \\
		& Time (s)
		& 243.6 & 289.0 & 335.3 & 293.0
		& 243.6 & 276.9 & 320.8 & 267.5 \\
		\addlinespace[2pt]
		\multirow{2}{*}{Byz-AdaVR-EF21} & Volume (MB)
		& 103.57 & 47.08 & $>$470.78 & 47.08
		& 47.08 & 47.08 & $>$470.78 & 42.37 \\
		& Time (s)
		& 436.9 & 197.0 & $>$2106.0 & 226.4
		& 198.6 & 197.0 & $>$2106.0 & 203.8 \\
		\addlinespace[2pt]
		\multirow{2}{*}{C-RSA} & Volume (MB)
		& \textbf{15.07} & 18.83 & \textbf{15.07} & \textbf{15.07}
		& \textbf{15.07} & $>$376.63 & \textbf{15.07} & 15.07 \\
		& Time (s)
		& \textbf{69.3} & 92.4 & 85.5 & 84.4
		& 69.3 & $>$1848.6 & 85.5 & \textbf{84.4} \\
		\addlinespace[2pt]
		\multirow{2}{*}{RAGE-Local-SGD} & Volume (MB)
		& \textbf{15.07} & \textbf{14.12} & \textbf{15.07} & \textbf{15.07}
		& \textbf{15.07} & \textbf{14.12} & \textbf{15.07} & \textbf{14.12} \\
		& Time (s)
		& 279.9 & 282.5 & 360.3 & 343.9
		& 279.9 & 282.5 & 360.3 & 322.4 \\
		\midrule
		\multirow{2}{*}{\textbf{Byrd-NAFL (ours)}} & Volume (MB)
		& 23.54 & 23.54 & 23.54 & 18.83
		& 18.83 & 18.83 & 23.54 & 23.54 \\
		& Time (s)
		& 84.1 & \textbf{87.2} & \textbf{81.0} & \textbf{76.1}
		& \textbf{67.3} & \textbf{69.8} & \textbf{81.0} & 95.1 \\
		\bottomrule
	\end{tabular}
\end{table*}

\subsubsection{\bfseries Malicious Attacks}
We evaluate the Byzantine resilience of Byrd-NAFL under four attacks.
Fall of Empires (FOE)~\cite{xie2019fallempiresbreakingbyzantinetolerant}, Label-Flipping (LF)~\cite{sf}, and SF~\cite{sf} are applied in both the i.i.d. and non-i.i.d. setups, and Mimic (MC)~\cite{mimic} is applied in the non-i.i.d. setup.
The no-attack (NA) case serves as a reference.
FOE, SF, and MC are omniscient attacks that use the honest gradients of the current iteration, whereas LF only manipulates the labels of the local data of the Byzantine workers.

In the \textbf{FOE} attack, the Byzantine workers send the average of the honest gradients scaled by a negative factor $-\mu$, i.e.,
\begin{equation}\label{eq:foe}
	g^*_{n,k} = -\frac{\mu}{H} \sum_{h=1}^{H} g_{h,k}, \quad n = H+1, \ldots, N,
\end{equation}
where we set $\mu = 10$.

In the \textbf{LF} attack, each Byzantine worker maps every label $y \in \{0, \ldots, C-1\}$ in its mini-batch $\zeta_{n,k}$ to $C-1-y$, where $C$ is the number of classes, and sends the gradient computed on the relabeled mini-batch $\zeta'_{n,k}$ as
\begin{equation}\label{eq:lf}
	g^*_{n,k} = \nabla \ell(x_k; \zeta'_{n,k}), \quad n = H+1, \ldots, N.
\end{equation}

In the \textbf{SF} attack, the Byzantine workers send the negated average of the honest gradients as
\begin{equation}\label{eq:sf}
	g^*_{n,k} = -\frac{1}{H} \sum_{h=1}^{H} g_{h,k}, \quad n = H+1, \ldots, N.
\end{equation}

In the \textbf{MC} attack, the Byzantine workers exploit the data heterogeneity in the non-i.i.d. setup.
They estimate the direction of maximum variance of the honest gradients (i.e., their first principal component), identify the honest worker $i^* \in \{1, \ldots, H\}$ whose gradient has the largest projection onto this direction, and replicate its gradient as
\begin{equation}\label{eq:mc}
	g^*_{n,k} = g_{i^*,k}, \quad n = H+1, \ldots, N.
\end{equation}

\begin{figure*}[t]
	\centering
	\includegraphics[width=0.95\linewidth]{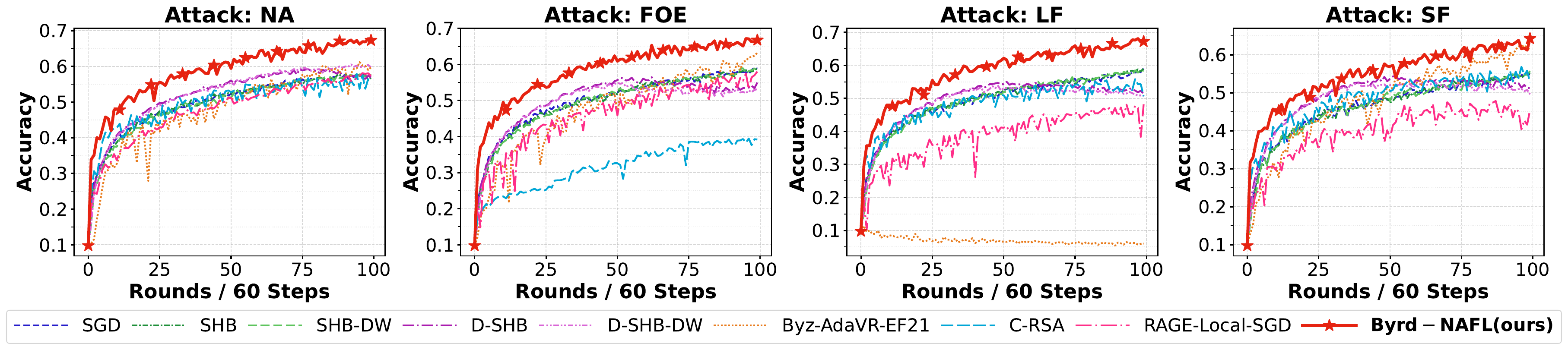}
	\caption{Test accuracy in the i.i.d. setup ($\epsilon=0.2$) for the CIFAR-10 dataset.}
	\label{fig:cifar_opt_02}
	\includegraphics[width=0.95\linewidth]{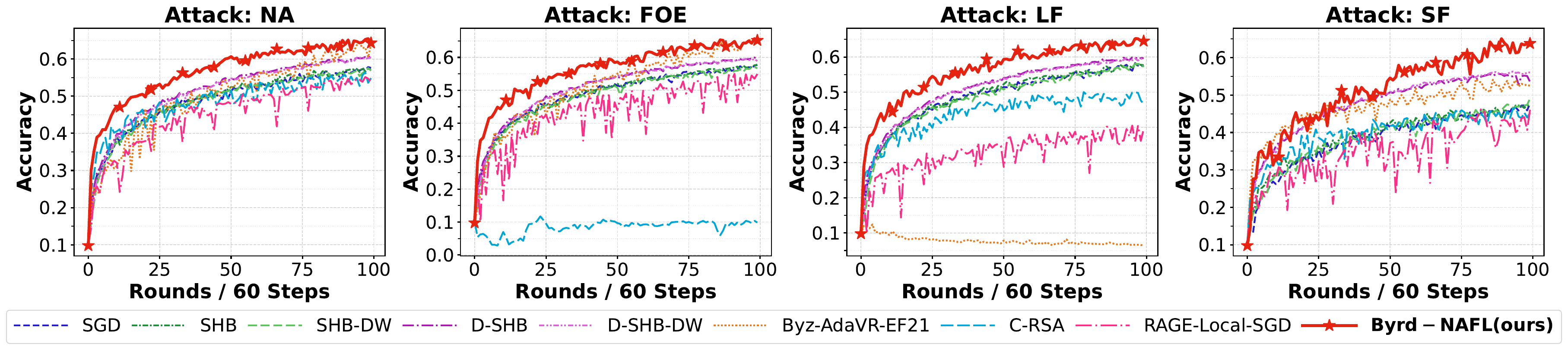}
	\caption{Test accuracy in the i.i.d. setup ($\epsilon=0.3$) for the CIFAR-10 dataset.} 
	\label{fig:cifar_opt_03}
	\includegraphics[width=0.95\linewidth]{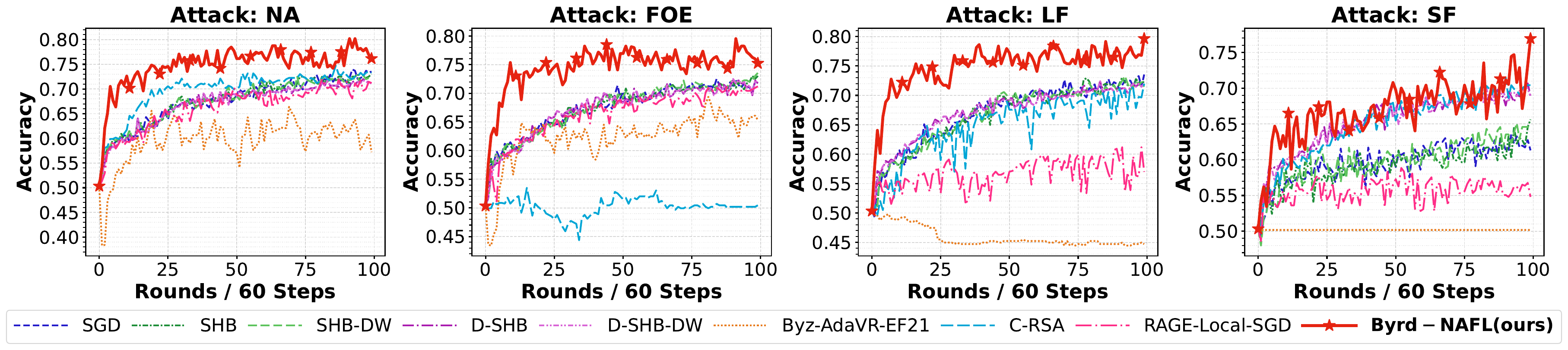}
	\caption{Test accuracy in the i.i.d. setup ($\epsilon=0.4$) for the Sentiment140 dataset.} 
	\label{fig:sent_opt_0.4}
\end{figure*}

\begin{table*}[t]
	\centering
	\caption{Per-worker communication volume and wall-clock time to reach 80\% of the peak accuracy on CIFAR-10. $>$: target not reached within 6,000 steps; bold: best value per column and metric.}
	\label{tab:communication_cifar}
	\begin{tabular}{lc cccc cccc}
		\toprule
		\multirow{2}{*}{Method} & \multirow{2}{*}{Metric}
		& \multicolumn{4}{c}{CIFAR-10 ($\epsilon = 0.2$)}
		& \multicolumn{4}{c}{CIFAR-10 ($\epsilon = 0.3$)} \\
		\cmidrule(lr){3-6} \cmidrule(lr){7-10}
		& & NA & FOE & LF & SF & NA & FOE & LF & SF \\
		\midrule
		\multicolumn{2}{l}{Target Accuracy}
		& 54\% & 54\% & 54\% & 51\% & 52\% & 53\% & 52\% & 52\% \\
		\midrule
		\multirow{2}{*}{SGD} & Volume (GB)
		& 162.51 & 157.27 & 180.86 & 175.62
		& 141.54 & 157.27 & 146.78 & $>$262.11 \\
		& Time (s)
		& 1313.2 & 1400.4 & 2078.3 & 1648.2
		& 1143.7 & 1400.4 & 1686.7 & $>$2460.0 \\
		\addlinespace[1pt]
		\multirow{2}{*}{SHB} & Volume (GB)
		& 162.51 & 152.03 & 167.75 & 172.99
		& 133.68 & 146.78 & 133.68 & $>$262.11 \\
		& Time (s)
		& 870.5 & 998.8 & 1743.4 & 1469.2
		& 716.0 & 964.3 & 1389.2 & $>$2226.0 \\
		\addlinespace[1pt]
		\multirow{2}{*}{SHB-DW} & Volume (GB)
		& 162.51 & 157.27 & 165.13 & 175.62
		& 141.54 & 149.40 & 138.92 & $>$262.11 \\
		& Time (s)
		& 1558.7 & 1666.8 & 1625.4 & 1632.1
		& 1357.6 & 1583.5 & 1367.4 & $>$2436.0 \\
		\addlinespace[1pt]
		\multirow{2}{*}{D-SHB} & Volume (GB)
		& 110.09 & 102.22 & 125.81 & 89.12
		& 107.47 & 110.09 & 110.09 & 157.27 \\
		& Time (s)
		& 1363.3 & 1345.5 & 2249.3 & 1379.0
		& 1330.9 & 1449.0 & 1968.1 & 2433.6 \\
		\addlinespace[1pt]
		\multirow{2}{*}{D-SHB-DW} & Volume (GB)
		& 115.33 & 112.71 & $>$262.11 & 91.74
		& 115.33 & 115.33 & 112.71 & 154.65 \\
		& Time (s)
		& 1401.8 & 1465.4 & $>$4320.0 & 1467.9
		& 1401.8 & 1499.5 & 1857.6 & 2474.5 \\
		\addlinespace[1pt]
		\multirow{2}{*}{Byz-AdaVR-EF21} & Volume (GB)
		& 154.65 & 157.27 & $>$262.11 & 136.30
		& 112.71 & 104.85 & $>$262.11 & 199.21 \\
		& Time (s)
		& 2757.7 & 2905.2 & $>$4758.0 & 3157.4
		& 2009.8 & 1936.8 & $>$4758.0 & 4614.7 \\
		\addlinespace[1pt]
		\multirow{2}{*}{C-RSA} & Volume (GB)
		& 73.39 & $>$131.06 & 96.98 & 65.53
		& 74.70 & $>$131.06 & $>$131.06 & $>$131.06 \\
		& Time (s)
		& 1256.6 & $>$2568.0 & 2908.2 & 2253.0
		& 1279.1 & $>$2568.0 & $>$3930.0 & $>$4506.0 \\
		\addlinespace[1pt]
		\multirow{2}{*}{RAGE-Local-SGD} & Volume (GB)
		& \textbf{38.27} & \textbf{38.79} & $>$52.42 & $>$52.42
		& \textbf{40.37} & \textbf{38.79} & $>$52.42 & $>$52.42 \\
		& Time (s)
		& 1778.3 & 2029.1 & $>$3624.0 & $>$4122.0
		& 1875.7 & 2029.1 & $>$3624.0 & $>$4122.0 \\
		\midrule
		\multirow{2}{*}{\textbf{Byrd-NAFL (ours)}} & Volume (GB)
		& 60.29 & 57.66 & \textbf{68.15} & \textbf{62.91}
		& 57.66 & 60.29 & \textbf{65.53} & \textbf{102.22} \\
		& Time (s)
		& \textbf{571.3} & \textbf{583.4} & \textbf{881.4} & \textbf{898.6}
		& \textbf{546.5} & \textbf{610.0} & \textbf{847.5} & \textbf{1460.2} \\
		\bottomrule
	\end{tabular}
\end{table*}

\subsection{Performance Comparison in the I.I.D. Setup}
\label{sec:4.B}
In the i.i.d. setup, we compare Byrd-NAFL with eight baselines: 
vanilla SGD, 
the server-side heavy-ball momentum methods (SHB~\cite{shb} and SHB-DW~\cite{shbdw}), 
the worker-side momentum methods (D-SHB~\cite{allouah2023fixingmixingrecipeoptimal,el2021distributed,Farhadkhani2022Resam}, D-SHB-DW), 
Byz-AdaVR-EF21~\cite{byz-adavr} with Adam-type momentum, 
a gradient-compression based method (C-RSA~\cite{crsa}),
and a communication frequency reduction method (RAGE-Local-SGD~\cite{rang}).

\subsubsection{\bfseries MNIST}
Figures~\ref{fig:opt_03} and~\ref{fig:opt_04} show the test accuracy on MNIST at $\epsilon=0.3$ and $\epsilon=0.4$, respectively.
Byrd-NAFL converges as fast as C-RSA and much faster than the other baselines, and it ranks first or second in final accuracy in all eight settings.
C-RSA slightly outperforms Byrd-NAFL in several settings but is less robust: at $\epsilon=0.4$, its accuracy peaks at 64.95\% under FOE, 29.32\% below that of Byrd-NAFL, and drops late in training under LF.
C-RSA also degrades on CIFAR-10 and Sentiment140, as shown below.
RAGE-Local-SGD reaches a final accuracy similar to that of Byrd-NAFL (e.g., 95.74\% vs.~94.98\% under FOE at $\epsilon=0.3$ and 92.77\% vs.~93.21\% under SF at $\epsilon=0.4$), but its early-stage curves almost overlap with those of SGD.
Byrd-NAFL outperforms all momentum-based baselines, e.g., the runner-up SHB-DW by 3.14\% under FOE and SHB by 3.10\% under SF at $\epsilon=0.3$, and SHB by at least 2.64\% at $\epsilon=0.4$, while Byz-AdaVR-EF21 fails under LF.
Hence, Byrd-NAFL is the only method that is both fast and robust in all MNIST settings.

Table~\ref{tab:communication_mnist} reports the per-worker communication volume and wall-clock time required to reach 80\% of the peak accuracy.
C-RSA and RAGE-Local-SGD need far less communication than the other baselines, e.g., 15.07~megabytes (MB) under SF at $\epsilon=0.3$, which is 68.0\%--86.1\% less, although C-RSA never reaches the target under FOE at $\epsilon=0.4$.
Byrd-NAFL needs 18.83--23.54~MB, slightly more than these two methods, but it has the shortest wall-clock time in six of the eight scenarios and the second shortest, after C-RSA, in the other two.
For instance, under SF at $\epsilon=0.3$, it reaches the target within 76.1~s, i.e., 9.8\% and 77.9\% less than C-RSA and RAGE-Local-SGD, respectively.
Since Byrd-NAFL neither compresses nor skips communication, these savings come from reaching the target accuracy in fewer communication steps.

\begin{figure*}[ht]
	\centering
	\includegraphics[width=0.95\linewidth]{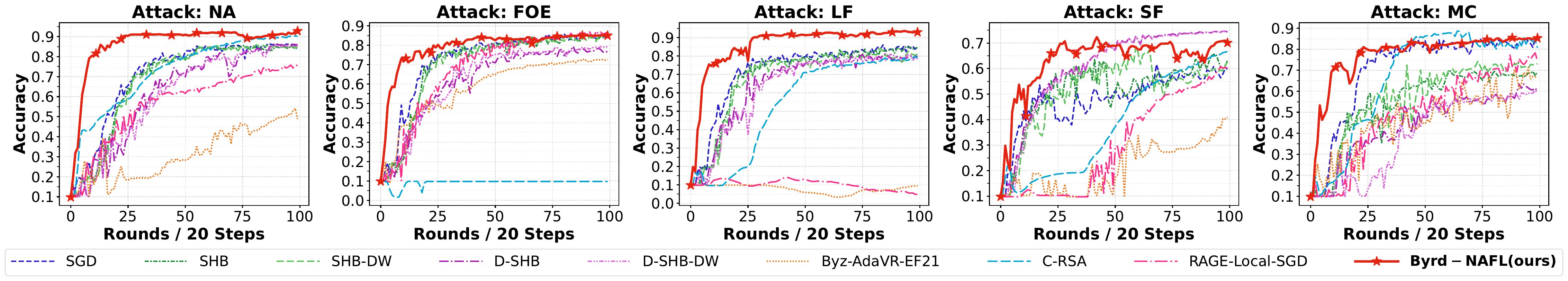}
	\caption{Test accuracy in the non-i.i.d. setup ($\epsilon=0.3$) for the MNIST dataset.} 
	\label{fig:noni.i.d.0.3}
	\includegraphics[width=0.95\linewidth]{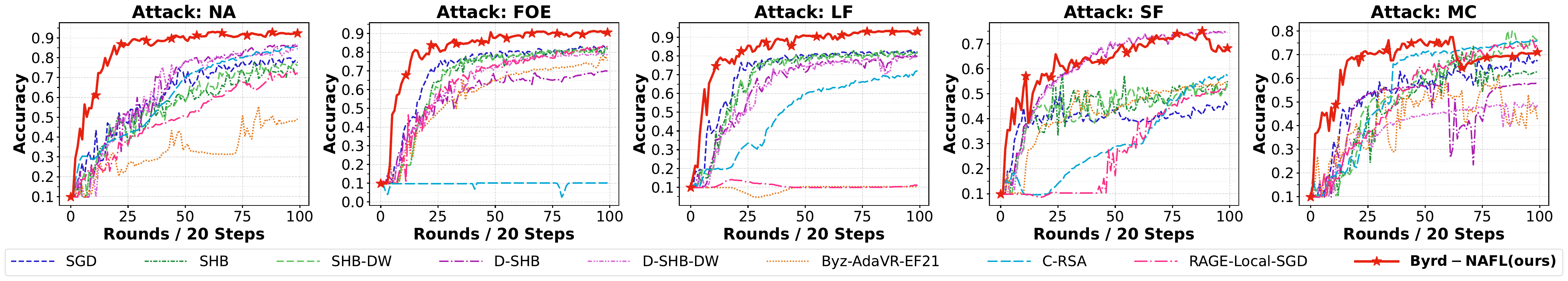}
	\caption{Test accuracy in the non-i.i.d. setup ($\epsilon=0.4$) for the MNIST dataset.} 
	\label{fig:noni.i.d.0.4}
	\includegraphics[width=0.95\linewidth]{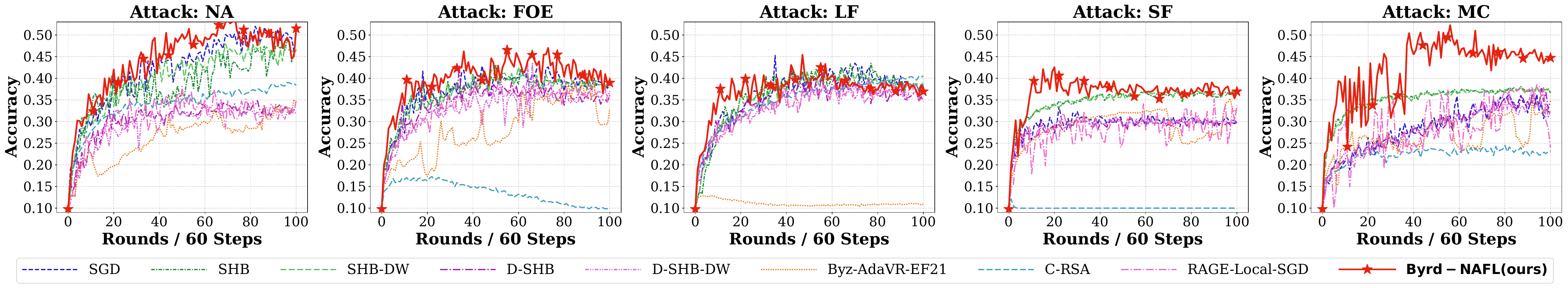}
	\caption{Test accuracy in the non-i.i.d. setup ($\epsilon=0.2$) for the CIFAR-10 dataset.} 
	\label{fig:cifar10_noniid_0.2}
	\includegraphics[width=0.95\linewidth]{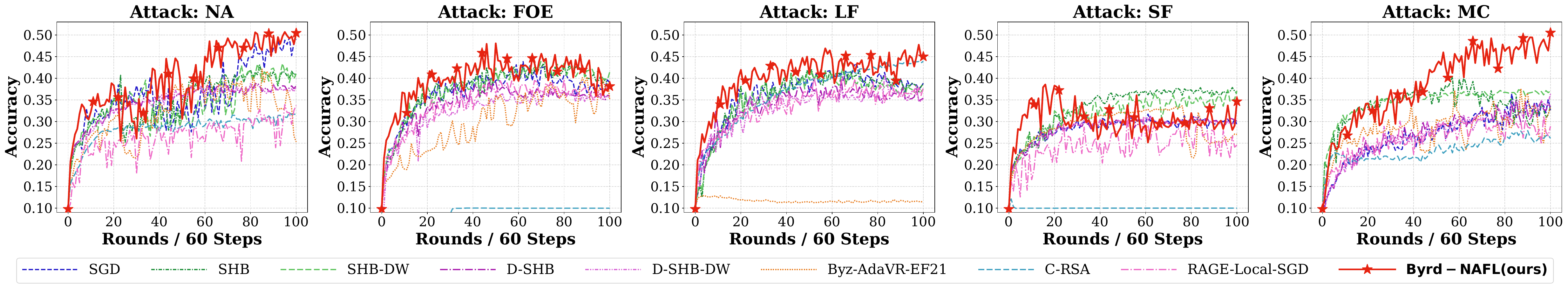}
	\caption{Test accuracy in the non-i.i.d. setup ($\epsilon=0.3$) for the CIFAR-10 dataset.} 
	\label{fig:cifar10_noniid_0.3}
	\includegraphics[width=0.95\linewidth]{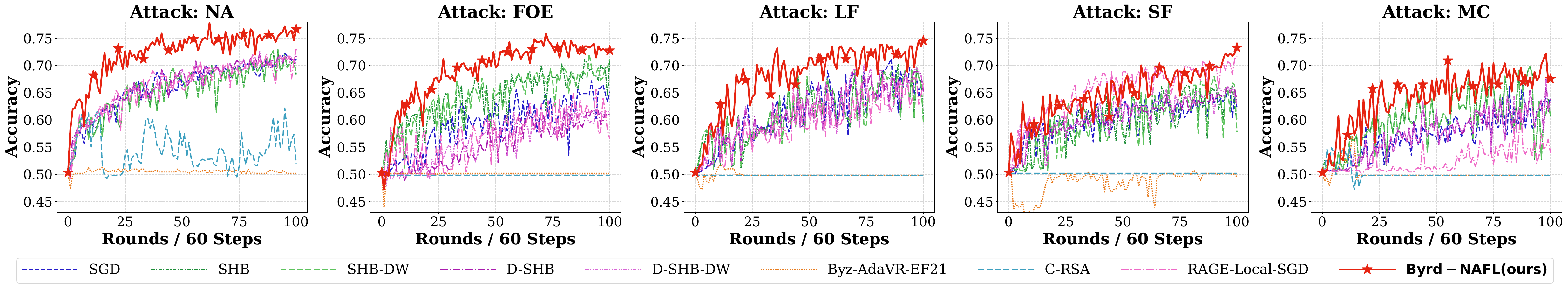}
	\caption{Test accuracy in the non-i.i.d. setup ($\epsilon=0.3$) for the Sentiment140 dataset.} 
	\label{fig:sentiment_transformer_noniid}
\end{figure*}

\subsubsection{\bfseries CIFAR-10}
\label{sec:cifa10}
Figures~\ref{fig:cifar_opt_02} and~\ref{fig:cifar_opt_03} show the test accuracy on CIFAR-10 at $\epsilon=0.2$ and $\epsilon=0.3$, respectively.
Byrd-NAFL converges fastest and attains the highest peak accuracy in all settings, i.e., 64.26\%--68.11\% at $\epsilon=0.2$ and 65.10\%--65.70\% at $\epsilon=0.3$.
In particular, LF barely affects Byrd-NAFL: its accuracy under LF is within 0.27\% of that under NA at $\epsilon=0.3$ and even slightly exceeds it at $\epsilon=0.2$ (68.11\% vs.~67.77\%).
Among the baselines, the worker-side momentum methods (D-SHB, D-SHB-DW, and Byz-AdaVR-EF21) are more resilient than SGD, SHB, and SHB-DW.
Byz-AdaVR-EF21 comes closest to Byrd-NAFL under SF at $\epsilon=0.2$ (62.48\%) and FOE at $\epsilon=0.3$ (65.28\%) but fails under LF at both ratios (11.42\% and 12.42\%).
C-RSA and RAGE-Local-SGD degrade more than on MNIST: C-RSA reaches only 11.73\% under FOE at $\epsilon=0.3$, and RAGE-Local-SGD stays between 40.53\% and 48.23\% under LF and SF.

Table~\ref{tab:communication_cifar} reports the communication volume and wall-clock time required to reach 80\% of the peak accuracy.
C-RSA and RAGE-Local-SGD need less communication than the other baselines, but within 6,000 steps C-RSA never reaches the target under FOE and RAGE-Local-SGD never reaches it under LF and SF, at either Byzantine ratio.
Even when they reach the target, they can be slower than SGD, e.g., by 44.9\% for RAGE-Local-SGD under FOE at $\epsilon=0.3$ and by 39.9\% for C-RSA under LF at $\epsilon=0.2$.
In contrast, Byrd-NAFL needs less communication than C-RSA in all eight scenarios, achieving the lowest volume under LF and SF and the second lowest, after RAGE-Local-SGD, under NA and FOE.
It also has the shortest wall-clock time in all scenarios, at least 23.7\% shorter than that of the second-fastest method.
Hence, on ResNet-18, faster convergence of Byrd-NAFL reduces the number of communication steps, which lowers both the communication volume and the wall-clock time while preserving Byzantine resilience.

\subsubsection{\bfseries Sentiment140}
\label{sec:sent}
We further evaluate Byrd-NAFL on the natural language processing (NLP) task of Sentiment140 with SentimentTransformer, whose attention-based sequence modeling typically requires higher optimization stability than image classification.
As shown in Figure~\ref{fig:sent_opt_0.4}, Byrd-NAFL leads all baselines by a clear margin under NA, FOE, and LF and attains the highest accuracy under SF, reaching 80.21\%, 79.54\%, and 76.93\% under NA, FOE, and SF, respectively, whereas the best baselines stay around 73\%.
Among the baselines, Byz-AdaVR-EF21 fails under LF and SF (50.33\%), C-RSA stays below 55\% under FOE, and RAGE-Local-SGD stays below 60\% under SF.
Moreover, the accuracy of Byrd-NAFL rises much faster than that of all baselines.
Hence, the convergence speed and Byzantine resilience of Byrd-NAFL carry over to this NLP task with a Transformer-based model.

\begin{figure*}[t]
	\centering
	\includegraphics[width=0.95\linewidth]{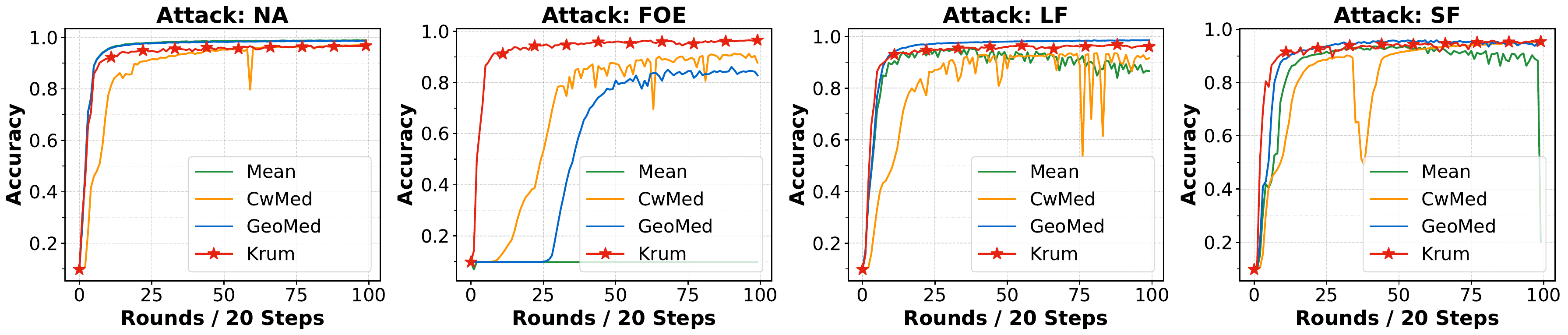}
	\caption{Ablation study on aggregation rules for the MNIST dataset.}
	\label{fig:MNIST_agg_ablation}
	\includegraphics[width=0.95\linewidth]{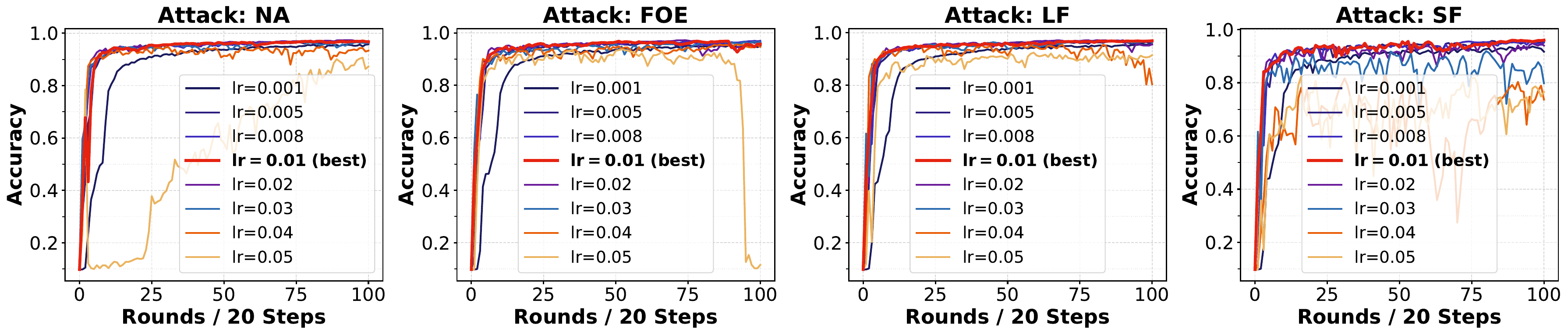}
	\caption{Ablation study on stepsize for the MNIST dataset.}
	\label{fig:MNIST-lr-ablation} 
	\includegraphics[width=0.95\linewidth]{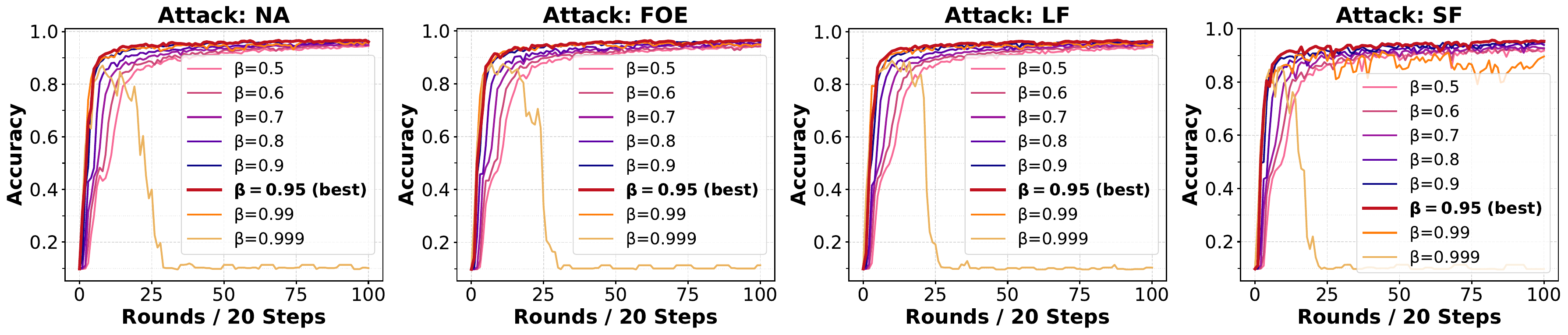}
	\caption{Ablation study on momentum for the MNIST dataset.}
	\label{fig:MNIST-beta-ablation}
\end{figure*}

\subsection{Performance Comparison in the Non-I.I.D. Setup}
\label{sec:4.C}
To reflect the data heterogeneity of real-world FL, we perform the comparisons in the non-i.i.d. setup.
Each training set is partitioned among the workers according to a Dirichlet distribution with concentration parameter $\alpha = 0.1$, $0.5$, and $0.5$ for MNIST, CIFAR-10, and Sentiment140, respectively, where a smaller $\alpha$ yields stronger heterogeneity.
In addition to FOE, LF, and SF, we also apply the MC attack that is designed to exploit data heterogeneity.

\subsubsection{\bfseries MNIST}
Figures~\ref{fig:noni.i.d.0.3} and~\ref{fig:noni.i.d.0.4} show the test accuracy on MNIST at $\epsilon=0.3$ and $\epsilon=0.4$, respectively.
The baselines perform inconsistently.
Among the worker-side momentum methods, D-SHB-DW outperforms Byrd-NAFL under SF at $\epsilon=0.3$, but D-SHB and D-SHB-DW degrade severely under MC, and Byz-AdaVR-EF21 performs poorly in most settings, which is consistent with the drift of local momentum vectors under data heterogeneity~\cite{allouah2023fixingmixingrecipeoptimal}.
Among the other baselines, SHB-DW outperforms Byrd-NAFL under MC at $\epsilon=0.4$, whereas SHB drops sharply under SF at $\epsilon=0.4$, C-RSA fails under FOE, and RAGE-Local-SGD fails under LF.
In contrast, Byrd-NAFL attains the highest or second-highest accuracy in all settings and converges fastest in almost all of them, remaining robust under strong data heterogeneity.

\subsubsection{\bfseries CIFAR-10}
\label{sec:cifar_noniid}
Figures~\ref{fig:cifar10_noniid_0.2} and~\ref{fig:cifar10_noniid_0.3} show the test accuracy on CIFAR-10 at $\epsilon=0.2$ and $\epsilon=0.3$, respectively.
The trends are consistent with those on MNIST.
The worker-side momentum methods (D-SHB, D-SHB-DW, and Byz-AdaVR-EF21) are most affected by data heterogeneity, whereas the server-side momentum methods (SHB, SHB-DW, and Byrd-NAFL) generally achieve higher accuracy; e.g., under SF, SHB-DW exceeds D-SHB-DW by 5.91\% and 6.65\% at $\epsilon=0.2$ and $\epsilon=0.3$, respectively.
Moreover, C-RSA fails under FOE and SF, and Byz-AdaVR-EF21 fails under LF.
Byrd-NAFL is the most robust across attacks, exceeding the second-best method by 5.29\% and 0.71\% under SF and by 13.77\% and 10.17\% under MC at $\epsilon=0.2$ and $\epsilon=0.3$, respectively.

\subsubsection{\bfseries Sentiment140}
Figure~\ref{fig:sentiment_transformer_noniid} reports the results on Sentiment140 with $\epsilon = 0.3$.
Compared with the i.i.d. setup, data heterogeneity slows convergence and amplifies fluctuations, especially under attacks.
C-RSA and Byz-AdaVR-EF21 fail to converge even under NA and remain at or below 50\% under all attacks.
Despite the fluctuations, Byrd-NAFL attains the highest accuracy in all settings, reaching 75.89\%, 74.59\%, 73.29\%, and 72.25\% under FOE, LF, SF, and MC, respectively, and generally converges fastest.
These results indicate that Byrd-NAFL remains effective in the heterogeneous setup with a Transformer-based model.

\begin{figure*}[tb]
	\centering
	\includegraphics[width=0.95\linewidth]{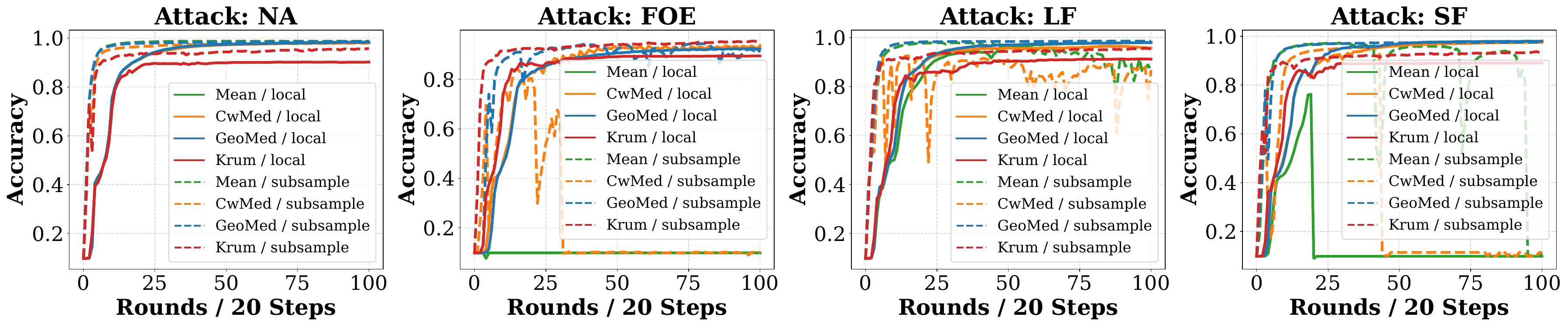}
	\caption{Test accuracy of Byrd-NAFL under partial worker participation and multiple local steps on the MNIST dataset.} 
	\label{fig:client_local}
	\includegraphics[width=0.95\linewidth]{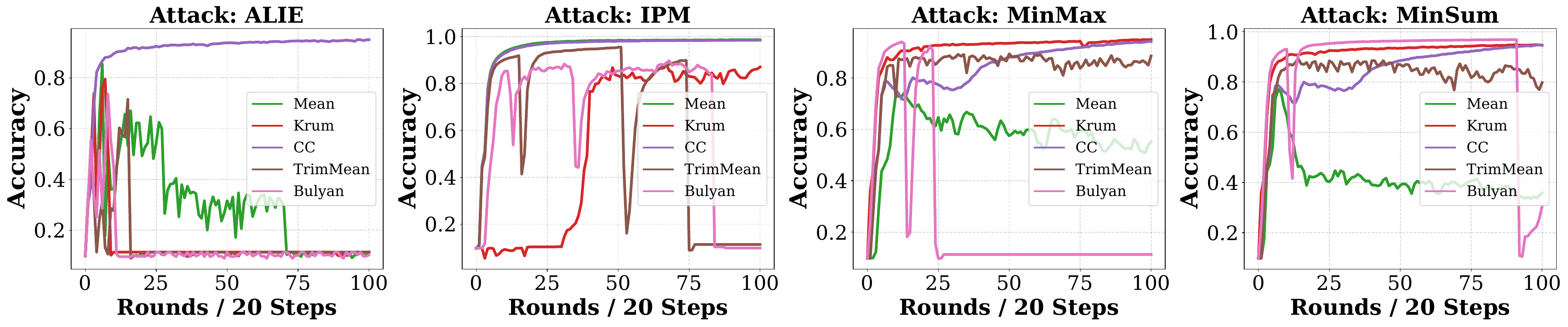}
	\caption{Test accuracy comparison under additional attacks and aggregation rules on the MNIST dataset.}
	\label{fig:additional_attacks_aggregations}
	\includegraphics[width=0.95\linewidth]{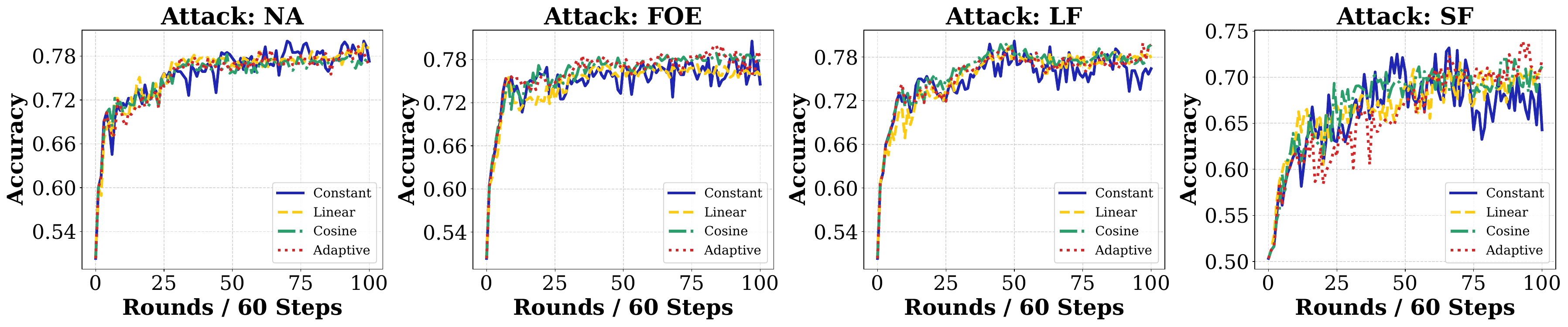}
	\caption{Test accuracy comparison of momentum decay strategies on the Sentiment140 dataset.}
	\label{fig:momentum_decay}
\end{figure*}

\subsection{Ablation Studies}
\label{sec:ablation}

\subsubsection{\bfseries Aggregation Rules Ablation}
\label{sec:agg_abl}
We run Byrd-NAFL with soft $\gamma$-Byzantine resilient rules (i.e., Krum, CwMed, GeoMed) and the non-robust Mean on MNIST with $\epsilon = 0.4$.
Figure~\ref{fig:MNIST_agg_ablation} presents the test accuracy under NA and the FOE, LF, and SF attacks.
Mean is not Byzantine-resilient: its accuracy stays at the chance level under FOE and collapses at the end of training under SF.
In contrast, all three robust rules withstand the three attacks; for instance, CwMed and GeoMed both exceed 80\% accuracy under the severe FOE attack.
{
This observation indicates that Byrd-NAFL is not tied to Krum and can be paired with other robust aggregation rules that satisfy Assumption~\ref{as:01}.
Nevertheless, CwMed and GeoMed converge more slowly than Krum, and CwMed shows occasional sharp accuracy drops.
Krum converges fast and remains stable under all three attacks, at the cost of a slightly lower accuracy than GeoMed under NA and LF. Moreover, different from GeoMed, Krum requires no iterative solver.
Based on these merits, we choose Krum as the default aggregation rule for our experiments in both the i.i.d. and non-i.i.d. setups.}

\subsubsection{\bfseries  Stepsize Ablation}
We investigate the impacts of the stepsize $\eta$ on Byrd-NAFL on MNIST with $\epsilon=0.4$ and $\beta=0.95$ by selecting $\eta \in \{0.001, 0.005, 0.008, 0.01,\allowbreak 0.02, 0.03, 0.04, 0.05\}$ around the tuned value $\eta=0.01$.
As shown in Figure~\ref{fig:MNIST-lr-ablation}, the stepsize $\eta=0.001$ induces a slower convergence of Byrd-NAFL with a similar final accuracy, whereas increasing the stepsize induces oscillations.
Moreover, Byrd-NAFL even collapses when $\eta=0.05$, and $\eta \ge 0.03$ clearly lowers the accuracy under SF.
{
Overall, stepsizes between 0.01 and 0.02 perform best.
This agrees with Theorem~\ref{thm:01}, whose initialization-dependent term decreases and whose error floor increases with $\eta$, both scaled by $1/\rho$: a small $\eta$ slows the early-stage convergence, whereas a large $\eta$ enlarges the error floor, especially under strong attacks.}

\subsubsection{\bfseries Momentum Ablation}
We vary the momentum factor $\beta \in \{0.5, 0.6, 0.7, 0.8, 0.9, 0.95, 0.99, 0.999\}$ on MNIST with $\eta=0.01$.
As shown in Figure~\ref{fig:MNIST-beta-ablation}, small momentum factors ($\beta \le 0.7$) slow down convergence, and $\beta=0.95$ performs best under all attacks.
Larger values degrade the performance: $\beta=0.99$ oscillates under SF, and $\beta=0.999$ collapses to the chance level in all settings after an initial rise.
{This observation is consistent with Theorem~\ref{thm:01}, whose sufficient stepsize bound scales with $\nicefrac{(1-\beta)^2}{(2\beta^2-\beta+1)}$: increasing $\beta$ from 0.95 to 0.99 tightens the bound by a factor of about 27, while the effective stepsize $\nicefrac{\eta}{(1-\beta)}$ grows from 0.2 to 1 at $\eta=0.01$.
	Hence, for a fixed $\eta$, a very large $\beta$ quickly leaves the regime covered by Theorem~\ref{thm:01}, and $\beta=0.95$ balances fast convergence against a moderate effective stepsize.}

\subsubsection{\bfseries Summary}
Byrd-NAFL is robust to moderate changes in $\eta$ and $\beta$ and performs best with $\eta=0.01$ and $\beta=0.95$, which we use as the default hyperparameters on MNIST.

\subsection{Generality Evaluation}
\label{sec:generality}
We next evaluate Byrd-NAFL under partial worker participation and multiple local steps, under additional attacks and aggregation rules, and with momentum decay.
Unless otherwise specified, the settings follow Section~\ref{sec:4.A}.

\subsubsection{\bfseries Partial Worker Participation and Multiple Local Steps}
IoT devices are often intermittently available and communicate infrequently.
We therefore consider two variants on MNIST: in partial participation, 50\% of the workers are randomly selected in each step, and in multiple local steps, each worker performs $\tau=5$ local updates between two communications (``subsample'' and ``local'' in Figure~\ref{fig:client_local}, respectively).
Each variant is combined with Mean, CwMed, GeoMed, and Krum.
As shown in Figure~\ref{fig:client_local}, Mean fails under FOE and SF in both variants, since it is not Byzantine-resilient.
Among the robust rules, only CwMed fails, and only with partial participation: it collapses under FOE and SF and oscillates under LF.
We attribute this to the step-to-step fluctuation of the Byzantine fraction among the selected workers, to which CwMed is more sensitive than GeoMed and Krum.
The other robust combinations do not collapse under any attack, although Krum with local steps saturates at about 90\% accuracy.
Hence, Byrd-NAFL works with both variants, and its stability under partial participation is determined by the aggregation rule.

\subsubsection{\bfseries Additional Attacks and Aggregation Rules}
On MNIST, we also apply four attacks designed to circumvent Byzantine-resilient aggregation, namely A Little Is Enough (ALIE)~\cite{baruch2019little}, Inner Product Manipulation (IPM)~\cite{xie2019fallempiresbreakingbyzantinetolerant}, and Min-Max and Min-Sum~\cite{shejwalkar2021manipulating}.
Besides Krum and the non-robust Mean, we combine Byrd-NAFL with Bulyan~\cite{mhamdi2018bulyan}, Centered Clipping (CC)~\cite{karimireddy2021learning}, and Trimmed Mean (TrimMean)~\cite{yin2018byzantine}.
As shown in Figure~\ref{fig:additional_attacks_aggregations}, only CC reaches high accuracy under all four attacks, albeit more slowly than Krum under Min-Max and Min-Sum.
Krum withstands Min-Max and Min-Sum, but it collapses under ALIE, and under IPM, it stays at the chance level for about 700 steps and then reaches only about 85\%.
TrimMean and Bulyan collapse under ALIE and IPM, and Bulyan also collapses under Min-Max and, near the end of training, under Min-Sum.
Mean withstands only IPM.
{
Hence, Byrd-NAFL inherits both the resilience and the vulnerabilities of its aggregation rule.
Consistent with Theorem~\ref{thm:01}, which gives no guarantee once the aggregation rule violates Assumption~\ref{as:01}, Nesterov's momentum does not prevent failures caused by the aggregation rule, whereas Byrd-NAFL converges reliably when paired with a suitable rule such as CC.}

\subsubsection{\bfseries Momentum Decay Strategies}
Since the training curves on Sentiment140 fluctuate noticeably, we examine whether decaying $\beta$ improves training stability.
Starting from $\beta_0=0.95$, linear and cosine decay reduce $\beta$ to $0$ over the training horizon, whereas adaptive decay sets $\beta_k=0.8\beta_{k-1}$ if $\big|\|\nabla_k\|-s_{k-1}\big|>0.9$ and $\beta_k=\beta_{k-1}$ otherwise, where $s_k=0.9s_{k-1}+0.1\|\nabla_k\|$ is a moving average of $\|\nabla_k\|$.
As shown in Figure~\ref{fig:momentum_decay}, all three strategies damp the late-stage oscillations of constant momentum and mostly raise the late-stage accuracy under attacks, most visibly under SF.
Cosine decay improves under all three attacks, whereas linear decay gains little under FOE, and adaptive decay fluctuates more in the middle of training under SF.
{
This trend agrees with Theorem~\ref{thm:01}: for a fixed $\eta$, the error floor scales with $\nicefrac{(2\beta^2-\beta+1)}{(1-\beta)^2}$, which increases with $\beta$, whereas the initialization-dependent term scales with $1-\beta$.
Hence, a large $\beta$ speeds up the early stage (cf. Figure~\ref{fig:MNIST-beta-ablation}), and decaying $\beta$ later lowers the error floor and stabilizes training.}

\section{Concluding Remarks}
\label{par:conclusion}
{We have proposed Byrd-NAFL that applies server-side Nesterov's momentum to the output of any soft $\gamma$-Byzantine-resilient aggregation rule to improve communication efficiency while retaining the resilience of the aggregation rule.}
For smooth non-convex losses, we have established an ${\cal O}\left(\nicefrac{(1-\beta)}{\eta K (1-\sin\gamma)}\right)$ convergence rate of Byrd-NAFL up to an error floor, which quantifies the impacts of Byzantine adversaries, Nesterov's momentum, and stochastic gradient noise.
Extensive experiments on MNIST, CIFAR-10, and Sentiment140 have shown that Byrd-NAFL outperforms the baselines in most i.i.d. and non-i.i.d. setups, although its resilience remains bounded by that of the underlying aggregation rule.
Since our evaluation is simulation-based and limited to moderate model sizes, future work will validate Byrd-NAFL on real heterogeneous edge devices with communication delays and hardware constraints, and extend it to larger models and more complex applications.

\appendices
{
\section{Proof of Theorem~\ref{thm:01}}
\label{app:proof-thm1}

Before starting the proof, we set $v_0 = x_0$, and let $\nabla_{k-1}$ denote the gradient obtained using the Byzantine-resilient aggregation rule at iteration $k-1$. 
We define the auxiliary residual-corrected sequence $v_k$ with $k\ge 1$ as
\begin{equation}\label{eqproof:01}
v_k  = \frac{1}{1-\beta}x_k -\frac{\beta}{1-\beta}x_{k-1} + \frac{\eta\beta}{1-\beta}\nabla_{k-1}
\end{equation}
where the correction term in $v_k$ explicitly captures the deviation from the nominal gradient-descent recursion. The sequence $v_k$ follows the auxiliary-sequence construction of~\cite{yang2016unified} for stochastic momentum methods; different from~\cite{yang2016unified}, our analysis requires neither unbiased nor bounded gradients and relies only on Assumptions~\ref{as:01} and~\ref{as:02}.

Based on \eqref{eqproof:01} and performing several algebraic manipulations, we have the difference between two consecutive $v_k$ as
\begin{equation}\label{eqproof:02}
    v_{k+1} - v_k = -\frac{\eta}{1-\beta} \nabla_k, k\ge 0
\end{equation}
and
\begin{equation}\label{eqproof:03}
    v_k - x_k = -\frac{\eta \beta^2 }{1-\beta} z_k.
\end{equation}

Note that the auxiliary sequence $v_k$ is introduced to decouple the momentum dynamics from the descent behavior of Byrd-NAFL to enable a tractable convergence analysis. 
While the original iterate $x_k$ evolves according to a momentum-based recursion that depends on historical gradients, the constructed sequence $v_k$ satisfies a simple gradient-descent-like update of the form $v_{k+1} = v_k - \frac{\eta}{1-\beta} \nabla_k$ as shown in \eqref{eqproof:02}. 
Such a transformation effectively removes the inertial effect induced by momentum and yields a virtual iterate that follows a clean descent trajectory. 
Hence, the decrease in the objective function can be analyzed along $v_k$ via standard smoothness arguments, which would be difficult when working directly with $x_k$. 
Moreover, the discrepancy between $v_k$ and $x_k$ captures the influence of momentum and can be explicitly bounded via the auxiliary variable $z_k$ that allows the error induced by historical accumulation to be isolated and controlled. 
Consequently, $v_k$ serves as a bridge that converts a complex and history-dependent optimization process into an equivalent form resembling gradient descent, while all additional effects from momentum and Byzantine perturbations are encapsulated into manageable error terms in the analysis.

Based on the recursion of $z_k$, we have 
\begin{equation}\label{eqproof:04}
    z_k = \sum_{t = 0}^{k-1} \beta^{k-1-t}  \nabla_t, k\ge 1
\end{equation}
where $z_0 = 0$.

By summing the squared norm of \eqref{eqproof:04} over $k=0, \ldots, K-1$ and recalling the fact that $z_0 = 0$, we obtain 
\begin{subequations}\label{eqproof:05}
    \begin{align}
        \sum_{k=0}^{K-1}\norm{ z_k }^2 &= \sum_{k=1}^{K-1}\NORM{  \sum_{t = 0}^{k-1} \beta^{k-1-t}  \nabla_t }^2  \\
        &= \sum_{k=1}^{K-1} w_k^2 \NORM{ \sum_{t = 0}^{k-1} \frac{\beta^{k-1-t}}{w_k}  \nabla_t }^2 \label{eqproof:05b} \\
        &\le \sum_{k=1}^{K-1} \sum_{t = 0}^{k-1} w_k \beta^{k-1-t} \norm{ \nabla_t }^2 \label{eqproof:05c}\\
        &\le \frac{1}{(1-\beta)^2} \sum_{k=0}^{K-1} \norm{ \nabla_k }^2 \label{eqproof:05d}
    \end{align}
\end{subequations}
where \eqref{eqproof:05b} is based on $w_k = \sum_{t = 0}^{k-1} \beta^{k-1-t} = \frac{ 1 - \beta^k }{1 - \beta}$; \eqref{eqproof:05c} follows from the convexity of the squared $\ell_2$-norm; and \eqref{eqproof:05d} is obtained by switching the ranges of subscript $k$ and $t$ and summing over $k$ with $\sum_{k=t+1}^{K-1} w_k \beta^{k-1-t} \le \frac{1}{(1-\beta)^2}$.

By using the $L$-Lipschitz continuous property, we have 
\begin{subequations}\label{eqproof:06}
    \begin{align}
        &f(v_{k+1}) - f(v_k) \notag\\
        &\le \inp{ \nabla f(v_k), v_{k+1} - v_k } 
        + \frac{L}{2}\norm{ v_{k+1} - v_k }^2 \label{eqproof:06a}\\
        &= \inp{ \nabla f(v_k) - \nabla f(x_k), v_{k+1} - v_k } \notag\\
        &\quad + \inp{ \nabla f(x_k), v_{k+1} - v_k } 
        + \frac{L}{2}\norm{ v_{k+1} - v_k }^2. \label{eqproof:06b}
    \end{align}
\end{subequations}

Substituting \eqref{eqproof:02} into \eqref{eqproof:06}, we have 
\begin{multline}\label{eqproof:07}
\inp{ \nabla f(x_k), \nabla_k } 
\le \inp{ \nabla f(x_k) - \nabla f(v_k), \nabla_k } \\
+ \frac{\eta L}{2(1-\beta)} \norm{ \nabla_k }^2
+ \frac{1 - \beta}{\eta}[f(v_k) - f(v_{k+1})]. 
\end{multline}

Summing the first term on the RHS of \eqref{eqproof:07} over $k=0,\ldots,K-1$, we have
\begin{subequations}\label{eqproof:08}
    \begin{align}
        &\sum_{k=0}^{K-1}\inp{ \nabla f(x_k) - \nabla f(v_k), \nabla_k } \notag\\
        &\le L\sum_{k=0}^{K-1}\norm{ x_k - v_k }\norm{ \nabla_k } \label{eqproof:08a} \\
        &= \frac{\eta L \beta^2}{1-\beta}\sum_{k=0}^{K-1}\norm{ z_k }\norm{ \nabla_k } \label{eqproof:08b} \\
        &\le \frac{\eta L \beta^2}{1-\beta}\sq{\sum_{k=0}^{K-1}\norm{ z_k }^2}^{\!\frac{1}{2}}\sq{\sum_{k=0}^{K-1}\norm{ \nabla_k }^2}^{\!\frac{1}{2}} \label{eqproof:08c}
    \end{align}
\end{subequations}
where \eqref{eqproof:08a} follows from the Cauchy--Schwarz inequality and the $L$-Lipschitz continuity of $\nabla f$; 
\eqref{eqproof:08b} follows from \eqref{eqproof:03}; and 
\eqref{eqproof:08c} follows from the Cauchy--Schwarz inequality.

Substituting \eqref{eqproof:05} into \eqref{eqproof:08c}, we obtain
\begin{equation}\label{eqproof:09}
\sum_{k=0}^{K-1}\inp{ \nabla f(x_k) - \nabla f(v_k), \nabla_k } 
\le \frac{\eta L \beta^2}{(1-\beta)^2}\sum_{k=0}^{K-1} \norm{\nabla_k}^2
\end{equation}
which holds for every $\beta\in[0,1)$; for $\beta=0$, $v_k=x_k$ and both sides of \eqref{eqproof:09} vanish.

Substituting \eqref{eqproof:09} into the summation of \eqref{eqproof:07} over $k=0, \ldots, K-1$, we obtain
\begin{equation}\label{eqproof:10}
\begin{split}
\sum_{k=0}^{K-1}\inp{ \nabla f(x_k), \nabla_k } 
&\le \frac{\eta L (2\beta^2 - \beta + 1)}{2(1-\beta)^2}
\sum_{k=0}^{K-1} \norm{\nabla_k}^2 \\
&\quad + \frac{1 - \beta}{\eta}[f(v_0) - f(v_{K})].
\end{split}
\end{equation}

Let $\mathcal{F}_k$ denote the $\sigma$-field generated by all randomness before the gradients of iteration $k$ are computed (in particular, by $x_0, \nabla_0, \ldots, \nabla_{k-1}$), and let $\eek{\cdot}$ denote the corresponding conditional expectation.
Hence, $x_k$, $z_k$, and $v_k$ are $\mathcal{F}_k$-measurable, whereas $\nabla_k$ can remain random conditional on $\mathcal{F}_k$.
Taking the expectation of the left-hand side of \eqref{eqproof:10}, based on the first condition of the soft $\gamma$-Byzantine resilience in Assumption~\ref{as:01}, we have 
\begin{equation}\label{eqproof:11}
\begin{split}
\eek{ \inp{ \nabla f(x_k), \nabla_k } } 
&= \inp{ \nabla f(x_k), \eek{\nabla_k} } \\
&\ge (1-\sin\gamma)\norm{\nabla f(x_k)}^2.
\end{split}
\end{equation}

Recalling the second condition of the soft $\gamma$-Byzantine resilience in Assumption~\ref{as:01}, we can recast the expectation of \eqref{eqproof:10} as
\begin{equation}\label{eqproof:12}
\begin{split}
&\sum_{k=0}^{K-1} \ee{ \norm{ \nabla f(x_k) }^2 } \\
&\le \frac{1 - \beta}{\eta(1-\sin\gamma)}\ee{ f(v_0) - f(v_{K}) } \\
&\quad + \frac{\eta L (2\beta^2 - \beta + 1)}{2(1-\beta)^2(1-\sin\gamma)} c_2 K \\
&\quad + \frac{\eta L c_1 (2\beta^2 - \beta + 1)}{2(1-\beta)^2(1-\sin\gamma)} \sum_{k=0}^{K-1}\ee{ \norm{ \nabla f(x_k) }^2 }.
\end{split}
\end{equation}

Since $v_0 = x_0$ and $f(v_K)\ge f^*$, we have $\ee{f(v_0)-f(v_K)}\le\Delta$; all expectations involved are finite, which follows by induction over $k$ from Assumptions~\ref{as:01} and~\ref{as:02} and $\ee{\norm{\nabla f(x_0)}^2}\le 2L\Delta$. 
With $\rho$ and $\phi_\beta$ defined in Theorem~\ref{thm:01}, the coefficient of the last term in \eqref{eqproof:12} is $\kappa := \nicefrac{\eta L c_1\phi_\beta}{(2(1-\beta)\rho)}$. 
Moving this term to the left-hand side and dividing both sides by $K(1-\kappa)$, we obtain, for any $\kappa<1$,
\begin{equation}\label{eqproof:13}
\frac{1}{K}\sum_{k=0}^{K-1} \ee{ \norm{\nabla f(x_k)}^2 } 
\le \frac{1-\beta}{1-\kappa}\Big[\frac{\Delta}{\rho \eta K} + \frac{L\phi_\beta c_2\eta}{2(1-\beta)^2\rho}\Big].
\end{equation}

The stepsize condition in Theorem~\ref{thm:01} is equivalent to $\kappa\le\nicefrac{1}{2}$, i.e., $\nicefrac{1}{(1-\kappa)}\le 2$, which yields \eqref{eq:18}.

\section{Proof of Theorem \ref{thm:02}}
\label{app:shb_cov}
The proof follows that of Theorem~\ref{thm:01} in Appendix~\ref{app:proof-thm1}, and we detail only the steps that differ.
In SHB, the server updates the global model as
\begin{equation}
        z_{k+1} = \beta z_k + \nabla_k, \quad  
        x_{k+1} = x_k - \eta z_{k+1}
    \label{eq_r2:hb-update}
\end{equation}
so that $x_k - x_{k-1} = -\eta z_k$ for $k\ge 0$, where $x_{-1} := x_0$ and $z_0 = 0$.
Analogously to~\eqref{eqproof:01}, but without the correction term, we define the auxiliary sequence $v_k = \frac{1}{1-\beta}x_k - \frac{\beta}{1-\beta}x_{k-1}$ for $k\ge 0$, so that $v_0 = x_0$.
Subtracting $x_k$ from $v_k$ and using $x_k - x_{k-1} = -\eta z_k$, we obtain the deviation
\begin{equation}
    v_k - x_k = \frac{\beta}{1-\beta}(x_k - x_{k-1}) = -\frac{\eta\beta}{1-\beta} z_k
    \label{eq_r2:hb-deviation}
\end{equation}
and, by the definition of $v_k$ and~\eqref{eq_r2:hb-update}
\begin{equation}
\begin{split}
    v_{k+1} - v_k &= \frac{(x_{k+1} - x_k) - \beta(x_k - x_{k-1})}{1-\beta} \\
    &= -\frac{\eta(z_{k+1} - \beta z_k)}{1-\beta} = -\frac{\eta}{1-\beta}\nabla_k
\end{split}
    \label{eq_r2:hb-aux-recursion}
\end{equation}
which coincides with~\eqref{eqproof:02}.

Hence, eqs. \eqref{eqproof:06} and~\eqref{eqproof:07} hold for the SHB iterates.
The only difference from Appendix~\ref{app:proof-thm1} lies in the deviation: by~\eqref{eq_r2:hb-deviation}, $\norm{x_k - v_k} = \frac{\eta\beta}{1-\beta}\norm{z_k}$ is proportional to $\beta$, whereas it is proportional to $\beta^2$ for Byrd-NAFL in~\eqref{eqproof:03}.
Replacing~\eqref{eqproof:03} by~\eqref{eq_r2:hb-deviation} in~\eqref{eqproof:08} and recalling~\eqref{eqproof:05}, we have
\begin{equation}
\begin{split}
&\sum_{k=0}^{K-1}\inp{\nabla f(x_k) - \nabla f(v_k), \nabla_k} \\
&\le \frac{\eta L\beta}{1-\beta}\sum_{k=0}^{K-1}\norm{z_k}\norm{\nabla_k}
\le \frac{\eta L\beta}{(1-\beta)^2}\sum_{k=0}^{K-1}\norm{\nabla_k}^2.
\end{split}
\label{eq_r2:hb-cross-sum}
\end{equation}

Substituting~\eqref{eq_r2:hb-cross-sum} into the summation of~\eqref{eqproof:07} over $k=0,\ldots,K-1$ and using $\frac{\eta L\beta}{(1-\beta)^2} + \frac{\eta L}{2(1-\beta)} = \frac{\eta L(1+\beta)}{2(1-\beta)^2} = \frac{\eta L\phi_\beta^{\mbox{\tiny HB}}}{2(1-\beta)}$, we obtain
\begin{equation}
\begin{split}
\sum_{k=0}^{K-1}\inp{\nabla f(x_k), \nabla_k}
&\le \frac{\eta L\phi_\beta^{\mbox{\tiny HB}}}{2(1-\beta)}\sum_{k=0}^{K-1} \norm{\nabla_k}^2 \\
&\quad + \frac{1-\beta}{\eta}\big[ f(v_0) - f(v_K) \big]
\end{split}
\label{eq_r2:hb-before-expect}
\end{equation}
which is~\eqref{eqproof:10} with $\phi_\beta = \nicefrac{(2\beta^2-\beta+1)}{(1-\beta)}$ replaced by $\phi_\beta^{\mbox{\tiny HB}}$.

Based on the arguments in~\eqref{eqproof:11}--\eqref{eqproof:13} with $\kappa$ replaced by $\kappa_{\mbox{\tiny HB}} := \nicefrac{\eta L c_1\phi_\beta^{\mbox{\tiny HB}}}{(2(1-\beta)\rho)}$ and setting $\eta \le \nicefrac{(1-\beta)\rho}{(L c_1\phi_\beta^{\mbox{\tiny HB}})}$, i.e., $\kappa_{\mbox{\tiny HB}}\le\nicefrac{1}{2}$, yields~\eqref{eq_r2:hb-final-bound1}.

\section{Proof of Proposition \ref{lem:equal-effective-stepsize}}\label{app:acc}
Unrolling~\eqref{eq:15} with $z_0=0$, as in~\eqref{eq:12}, gives the search direction of Byrd-NAFL as
\begin{equation}\label{r2_nafl_dir}
	\frac{1}{\eta}(x_k-x_{k+1}) = y_k = (1+\beta)\nabla_k+\beta^2\sum_{t=0}^{k-1}\beta^{k-1-t}\nabla_t
\end{equation}
and unrolling~\eqref{eq_r2:hb-update} gives that of SHB as
\begin{equation}\label{r2_shb_dir}
	\frac{1}{\eta}(x_k-x_{k+1}) = z_{k+1} = \nabla_k+\beta\sum_{t=0}^{k-1}\beta^{k-1-t}\nabla_t.
\end{equation}

Substituting $\nabla_t=g$ for $t=0,\ldots,k$ into~\eqref{r2_nafl_dir} and~\eqref{r2_shb_dir} and summing the geometric series gives $(x_k-x_{k+1})/\eta=(1-\beta^{k+2})g/(1-\beta)$ for Byrd-NAFL and $(1-\beta^{k+1})g/(1-\beta)$ for SHB, i.e., the stated $\eta_{\mathrm{eff},k}$.
The difference between the two effective stepsizes is $\eta(\beta^{k+1}-\beta^{k+2})/(1-\beta)=\eta\beta^{k+1}$, and both tend to $\nicefrac{\eta}{(1-\beta)}$ as $k\to\infty$ since $0\le\beta<1$.
}

\bibliographystyle{IEEEtran}
\bibliography{ieee2026329}


\clearpage
\renewcommand{\theequation}{S\arabic{equation}}
\renewcommand{\thetable}{S\arabic{table}}
\renewcommand{\thefigure}{S\arabic{figure}}
\renewcommand{\thetheorem}{S\arabic{theorem}}
\renewcommand{\thelemma}{S\arabic{lemma}}
\renewcommand{\theassump}{S\arabic{assump}}
\renewcommand{\theremark}{S\arabic{remark}}
\renewcommand{\thecorollary}{S\arabic{corollary}}
\renewcommand{\theproposition}{S\arabic{proposition}}
\renewcommand{\thepage}{S\arabic{page}}

\onecolumn

\begin{center}
{\Large\bfseries Supplementary Materials\par}
\vspace{0.5em}
{\Large\bfseries
	A Byzantine-Robust Federated Learning with Nesterov's Momentum
	\par}
\end{center}
\vspace{1em}

\noindent
Appendix~\ref{app:constants} in supplementary materials follows Appendices~A--C of the main manuscript.
Equations, tables, lemmas, and corollaries numbered with the prefix~S, e.g., (S1), Table~S1, and Lemma~S1, belong to these supplementary materials, whereas all other numbered equations, assumptions, theorems, sections, figures, and references are those of the main manuscript.

\appendices
\setcounter{section}{3}

	\section{Byzantine-Resilient Hyperparameter Selection}
	\label{app:constants}
	Theorem~\ref{thm:01} depends on the Byzantine-resilient hyperparameters $(\gamma,c_1,c_2)$ in Assumption~\ref{as:01}, which are difficult to estimate in practice since they depend on the unknown true gradient, the data distribution, and the type and strength of the attack (Section~II-B).
	Conditions~\eqref{eq:07} and~\eqref{eq:08} concern only the aggregation step at the current iterate, i.e., $\eek{\nabla_k}$ and $\eek{\norm{\nabla_k}^2}$, which are determined by the aggregation rule, the attack, and the data at $x_k$, whereas the momentum affects them only through the trajectory $\{x_k\}$.
	Hence, $(\gamma,c_1,c_2)$ characterize the aggregation rule under a given attack and data distribution: they can be bounded in closed form for robust aggregation rules (Appendix~\ref{app:constants-closed}) and estimated in simulation for any aggregation rule (Appendix~\ref{app:constants-sim}).
	
	\subsection{Closed Form for Robust Aggregation Rules}
	\label{app:constants-closed}
	An aggregation rule is $(N-H,\lambda)$-robust~\cite{allouah2023fixingmixingrecipeoptimal} if, for any inputs $\tilde g_1,\ldots,\tilde g_N\in\mathbb{R}^d$ and any set $\mathcal{S}\subseteq\{1,\ldots,N\}$ with $|\mathcal{S}|=H$,
	\begin{equation}\label{eq:lambda-robust}
		\NORM{\bragg{[\tilde g_n]_{n=1}^N}-\bar g_{\mathcal{S}}}^2\le\frac{\lambda}{H}\sum_{n\in\mathcal{S}}\norm{\tilde g_n-\bar g_{\mathcal{S}}}^2
	\end{equation}
	where $\bar g_{\mathcal{S}}:=\frac{1}{H}\sum_{n\in\mathcal{S}}\tilde g_n$, and $\lambda$ is called the robustness coefficient (denoted by $\kappa$ in~\cite{allouah2023fixingmixingrecipeoptimal}).
	For instance, $\lambda=\nicefrac{6H}{(2H-N)}$ for Krum (in the variant analyzed in~\cite{allouah2023fixingmixingrecipeoptimal}), $\lambda=\nicefrac{4H^2}{(2H-N)^2}$ for GeoMed, and $\lambda=\nicefrac{6H(N-H)}{(2H-N)^2}$ for TrimMean that discards the $N-H$ largest and the $N-H$ smallest values per coordinate, whereas no aggregation rule achieves $\lambda<\nicefrac{(N-H)}{(2H-N)}$~\cite{allouah2023fixingmixingrecipeoptimal}.
	Table~\ref{tab:lambda} lists these coefficients for our settings.
	In contrast, CC with a fixed clipping radius $\tau_{\mathrm{c}}$ admits no finite $\lambda$: if the $H$ honest inputs are $0$ and the $N-H$ Byzantine inputs equal $\nicefrac{\tau_{\mathrm{c}}}{2}$ in one dimension, then no input is clipped, and CC initialized at $0$ returns $\nicefrac{(N-H)\tau_{\mathrm{c}}}{(2N)}\neq0$ after any number of iterations, although the right-hand side of~\eqref{eq:lambda-robust} vanishes.
	
	\begin{table}[!htbp]
		\centering
		\caption{Robustness coefficients $\lambda$ of $(N-H,\lambda)$-robust aggregation rules~\cite{allouah2023fixingmixingrecipeoptimal} in our settings.}
		\label{tab:lambda}
		\begin{tabular}{lcccccc}
			\toprule
			& \multicolumn{2}{c}{MNIST ($N=500$)} & \multicolumn{2}{c}{CIFAR-10 ($N=40$)} & \multicolumn{2}{c}{Sentiment140 ($N=20$)} \\
			\cmidrule(lr){2-3} \cmidrule(lr){4-5} \cmidrule(lr){6-7}
			Rule & $\epsilon=0.3$ & $\epsilon=0.4$ & $\epsilon=0.2$ & $\epsilon=0.3$ & $\epsilon=0.3$ & $\epsilon=0.4$ \\
			\midrule
			Krum & 10.5 & 18 & 8 & 10.5 & 10.5 & 18 \\
			GeoMed & 12.25 & 36 & 7.11 & 12.25 & 12.25 & 36 \\
			TrimMean & 7.88 & 36 & 2.67 & 7.88 & 7.88 & 36 \\
			Lower bound & 0.75 & 2 & 0.33 & 0.75 & 0.75 & 2 \\
			\bottomrule
		\end{tabular}
	\end{table}
	
	\begin{lemma}\label{lem:robust-constants}
		Suppose that, conditional on $\mathcal{F}_k$ (cf. Appendix~\ref{app:proof-thm1}), the honest stochastic gradients $g_{n,k}$, $n=1,\ldots,H$, are independent and unbiased with $\eek{\norm{g_{n,k}-\nabla f_n(x_k)}^2}\le\sigma^2$, and that $\frac{1}{H}\sum_{n=1}^{H}\norm{\nabla f_n(x)-\nabla f(x)}^2\le\omega^2$ for all $x$.
		Then, every $(N-H,\lambda)$-robust aggregation rule satisfies~\eqref{eq:08} with $c_1=2$ and $c_2=\nicefrac{2\sigma^2}{H}+2\lambda(\sigma^2+\omega^2)$, and
		\begin{equation}\label{eq:07r}
			\inp{\eek{\nabla_k},\nabla f(x_k)}\ge\frac{1}{2}\norm{\nabla f(x_k)}^2-c_3,\quad c_3:=\frac{\lambda(\sigma^2+\omega^2)}{2}.
		\end{equation}
		
		Moreover, for any $\nu>\sqrt{\lambda(\sigma^2+\omega^2)}$,~\eqref{eq:07} holds at every $x_k$ with $\norm{\nabla f(x_k)}\ge\nu$ for any $\gamma\in(0,\frac{\pi}{2})$ with $\sin\gamma\ge\nicefrac{\sqrt{\lambda(\sigma^2+\omega^2)}}{\nu}$.
	\end{lemma}
	
	\begin{IEEEproof}
		Let $\bar g_k:=\frac{1}{H}\sum_{n=1}^{H}g_{n,k}$, $e_k:=\nabla_k-\bar g_k$, and $V_k:=\frac{1}{H}\sum_{n=1}^{H}\norm{g_{n,k}-\bar g_k}^2$.
		By~\eqref{eq:lambda-robust} with $\mathcal{S}=\{1,\ldots,H\}$, we have $\norm{e_k}^2\le\lambda V_k$.
		Since $\bar g_k$ minimizes $\sum_{n=1}^{H}\norm{g_{n,k}-v}^2$ over $v\in\mathbb{R}^d$ and $\eek{g_{n,k}}=\nabla f_n(x_k)$, we have $\eek{V_k}\le\frac{1}{H}\sum_{n=1}^{H}\eek{\norm{g_{n,k}-\nabla f(x_k)}^2}\le\sigma^2+\omega^2$.
		Since $\eek{\bar g_k}=\nabla f(x_k)$, the Cauchy--Schwarz and Jensen inequalities give
		\begin{equation*}
			\inp{\eek{\nabla_k},\nabla f(x_k)}=\norm{\nabla f(x_k)}^2+\inp{\eek{e_k},\nabla f(x_k)}\ge\norm{\nabla f(x_k)}^2-\norm{\nabla f(x_k)}\sqrt{\lambda(\sigma^2+\omega^2)},
		\end{equation*}
		which yields~\eqref{eq:07} whenever $\norm{\nabla f(x_k)}\ge\nu$ and, by Young's inequality,~\eqref{eq:07r}.
		Moreover, $\eek{\norm{\nabla_k}^2}\le2\eek{\norm{\bar g_k}^2}+2\eek{\norm{e_k}^2}$ and, by the conditional independence, $\eek{\norm{\bar g_k}^2}\le\norm{\nabla f(x_k)}^2+\nicefrac{\sigma^2}{H}$, which yields~\eqref{eq:08} with the stated $c_1$ and $c_2$.
	\end{IEEEproof}
	
	\begin{corollary}\label{cor:relaxed}
		Suppose that Assumption~\ref{as:02} holds and that Assumption~\ref{as:01} holds with~\eqref{eq:07} replaced by $\inp{\eek{\nabla_k},\nabla f(x_k)}\ge\rho\norm{\nabla f(x_k)}^2-c_3$ for some $\rho\in(0,1]$ and $c_3\ge0$.
		Then, for $0\le\beta<1$ and under the stepsize condition of Theorem~\ref{thm:01}, the bound~\eqref{eq:18} holds with the additional term $\nicefrac{2c_3}{\rho}$ on its right-hand side.
		In particular, for any $(N-H,\lambda)$-robust aggregation rule under the conditions of Lemma~\ref{lem:robust-constants} and any $\eta\le\nicefrac{(1-\beta)}{(4L\phi_\beta)}$,
		\begin{equation}\label{eq:robust-bound}
			\frac{1}{K}\sum_{k=0}^{K-1}\ee{\norm{\nabla f(x_k)}^2}
			\le\frac{4(1-\beta)\Delta}{\eta K}+\frac{4L\phi_\beta\eta}{1-\beta}\Big(\frac{\sigma^2}{H}+\lambda(\sigma^2+\omega^2)\Big)+2\lambda(\sigma^2+\omega^2).
		\end{equation}
	\end{corollary}
	
	\begin{IEEEproof}
		Under the relaxed condition, the lower bound in~\eqref{eqproof:11} becomes $\rho\norm{\nabla f(x_k)}^2-c_3$, so that~\eqref{eqproof:12} holds with $1-\sin\gamma$ replaced by $\rho$ and the additional term $\nicefrac{c_3K}{\rho}$ on its right-hand side.
		Hence, the right-hand side of~\eqref{eqproof:13} gains $\nicefrac{c_3}{((1-\kappa)\rho)}\le\nicefrac{2c_3}{\rho}$.
		The bound~\eqref{eq:robust-bound} follows by substituting $\rho=\nicefrac{1}{2}$, $c_1=2$, and $c_2$ and $c_3$ from Lemma~\ref{lem:robust-constants}, for which the stepsize condition of Theorem~\ref{thm:01} reads $\eta\le\nicefrac{(1-\beta)}{(4L\phi_\beta)}$.
	\end{IEEEproof}
	
	\label{par:robust-implications}%
	The bound~\eqref{eq:robust-bound} also shows how the aggregation rule and the attack affect the selection of $\eta$ and $\beta$.
	First, it holds for every attack, and its stepsize condition depends only on $L$ and $\beta$, i.e., neither on the attack nor on the aggregation rule.
	Second, the aggregation rule and the Byzantine fraction affect the bound only through $\lambda$: a larger $\lambda$, i.e., a larger worst-case aggregation error, raises the error floor and, for a sufficiently large $K$, lowers the bound-optimal stepsize $\eta^\star=(1-\beta)\sqrt{\nicefrac{\Delta}{(L\phi_\beta K(\nicefrac{\sigma^2}{H}+\lambda(\sigma^2+\omega^2)))}}$, which supports calibrating $\eta$ where the aggregation error is largest, i.e., under the most severe anticipated attack, as in our tuning protocol (Section~\ref{sec:4.A}3); e.g., $\lambda$ of Krum increases from 10.5 to 18 when $\epsilon$ increases from 0.3 to 0.4 on MNIST (Table~\ref{tab:lambda}).
	Third, a larger honest mini-batch reduces $\sigma^2$ and thus both the error floor and the last term of~\eqref{eq:robust-bound}.
	Finally, the last term cannot be reduced by any choice of $\eta$ and $\beta$; its parts proportional to $\lambda\sigma^2$ and $\lambda\omega^2$ are consistent with the non-vanishing errors of memoryless robust aggregation of stochastic gradients~\cite{karimireddy2021learning} and of Byzantine-resilient learning under data heterogeneity~\cite{mimic}, respectively.
	Different from~\eqref{eq:07}, which Lemma~\ref{lem:robust-constants} guarantees only where the true gradient dominates the aggregation error, as also required for Krum in~\cite[Proposition~1]{blanchard2017machine}, the relaxed condition~\eqref{eq:07r} holds at every iterate.
	
	\subsection{Estimation in Simulation}
	\label{app:constants-sim}
	In simulation, the honest workers and their local datasets are known, so that~\eqref{eq:07} and~\eqref{eq:08} can be checked along a trajectory of Byrd-NAFL for any aggregation rule and attack.
	At checkpoints $k_1,\ldots,k_J$, we compute $\nabla f(x_k)$ from the full local datasets of the honest workers, freeze all $\mathcal{F}_k$-measurable quantities, e.g., the model and the internal states of the aggregation rule and of the attack, and replay iteration $k$ for $M$ times with fresh mini-batches and regenerated Byzantine inputs.
	This yields $M$ independent samples $\nabla_k^{(m)}$ of the aggregated gradient conditional on $\mathcal{F}_k$.
	Since $\nabla f(x_k)$ is exact, the sample means of $\nicefrac{\inp{\nabla_k^{(m)},\nabla f(x_k)}}{\norm{\nabla f(x_k)}^2}$ and $\norm{\nabla_k^{(m)}}^2$ are unbiased estimates of the alignment ratio $\rho_k:=\nicefrac{\inp{\eek{\nabla_k},\nabla f(x_k)}}{\norm{\nabla f(x_k)}^2}$ and of $\eek{\norm{\nabla_k}^2}$, respectively, and one-sided bootstrap confidence bounds at level $\nicefrac{\delta}{J}$ hold simultaneously for all checkpoints with probability at least $1-2\delta$ (asymptotically).
	Let $\underline{\rho}_j$ denote the lower confidence bound of $\rho_{k_j}$.
	Then, $1-\sin\gamma$ is estimated by $\min_j\underline{\rho}_j$, which certifies~\eqref{eq:07} at the checkpoints if it is positive, and $(c_1,c_2)$ by a line $c_1\norm{\nabla f(x_k)}^2+c_2$ above all upper confidence bounds of $\eek{\norm{\nabla_k}^2}$, e.g., the one that minimizes the sum of the vertical gaps, which is a linear program.
	If $\underline{\rho}_j$ deteriorates as $\norm{\nabla f(x_{k_j})}$ decreases, as Lemma~\ref{lem:robust-constants} allows, the relaxed condition~\eqref{eq:07r} is used instead: for a chosen $\rho$, e.g., $\rho=\nicefrac{1}{2}$, $c_3$ is estimated by $\max_j(\rho-\underline{\rho}_j)_+\norm{\nabla f(x_{k_j})}^2$.
	With $M=64$, the cost per checkpoint is $MH$ mini-batch gradients, e.g., 16 epochs of MNIST for $H=300$ and a mini-batch size of 50.

\end{document}